\documentclass[sigconf,nonacm]{acmart}

\setcopyright{none}
\renewcommand\footnotetextcopyrightpermission[1]{}

\usepackage{graphicx}
\usepackage{booktabs}
\usepackage{multirow}
\usepackage{makecell}
\usepackage{array}
\usepackage{placeins}
\usepackage{stfloats}
\newcolumntype{C}[1]{>{\centering\arraybackslash}m{#1}}
\usepackage{tikz}
\usetikzlibrary{arrows.meta,fit,positioning}
\title[FusionRS: RGB--Infrared-Style Remote Sensing Data]{FusionRS: A Large-Scale RGB--Infrared-Style Remote Sensing Dataset for Cross-Modal Vision--Language Learning}

\author{Jiaju Han}
\affiliation{%
  \institution{China University of Petroleum-Beijing at Karamay}
  \city{Karamay}
  \country{China}}
\additionalaffiliation{%
  \institution{Shenzhen Research Institute of Big Data}
  \city{Shenzhen}
  \country{China}}
\author{Ben Zhang}
\affiliation{%
  \institution{University of Electronic Science and Technology of China}
  \city{Chengdu}
  \country{China}}
\author{Xuemeng Sun}
\affiliation{%
  \institution{China University of Petroleum-Beijing at Karamay}
  \city{Karamay}
  \country{China}}
\author{Qike Zhang}
\affiliation{%
  \institution{China University of Petroleum-Beijing at Karamay}
  \city{Karamay}
  \country{China}}
\author{Yuxian Dong}
\affiliation{%
  \institution{China University of Petroleum-Beijing at Karamay}
  \city{Karamay}
  \country{China}}
\author{Dingyi Lu}
\affiliation{%
  \institution{China University of Petroleum-Beijing at Karamay}
  \city{Karamay}
  \country{China}}
\author{Chengyin Hu}
\affiliation{%
  \institution{China University of Petroleum-Beijing at Karamay}
  \city{Karamay}
  \country{China}}
\author{Luwei Yang}
\affiliation{%
  \institution{Shenzhen Research Institute of Big Data}
  \city{Shenzhen}
  \country{China}}
\author{Fengyu Zhang}
\affiliation{%
  \institution{China University of Petroleum-Beijing at Karamay}
  \city{Karamay}
  \country{China}}
\author{Yiwei Wei}
\affiliation{%
  \institution{Tianjin University}
  \city{Tianjin}
  \country{China}}
\author{Jiujiang Guo}
\affiliation{%
  \institution{North University of China}
  \city{Taiyuan}
  \country{China}}
\renewcommand{\shortauthors}{Han et al.}

\begin{document}

% Reduce spacing in author block
\settopmatter{authorsperrow=4}

\begin{abstract}
Remote sensing vision--language models have advanced Earth observation, but available large-scale vision--language resources remain RGB-centered, leaving complementary infrared information underexplored. Infrared observations provide distinctive intensity structures, object boundaries, and illumination-invariant cues that complement conventional RGB imagery, yet large-scale RGB--infrared--text resources remain scarce. We introduce FusionRS, the first large-scale RGB--infrared-style--text dataset for controlled dual-modal remote sensing vision--language learning. It contains 600,000 spatially aligned pairs created by translating diverse public RGB remote sensing images into infrared-style counterparts. Each pair retains a conventional scene caption, and a curated subset adds 45,913 IR-aware captions describing observable intensity, contrast, texture, and structure while preserving scene semantics. We train CLIP-style models for RGB--infrared-style--text alignment and adapt a generative vision--language model with mixed task-conditioned caption supervision. Evaluation covers cross-modal retrieval, scaling and supervision ablations, sensor-captured transfer, and strictly held-out captioning and VQA. FusionRS substantially improves RGB--infrared-style alignment and infrared-to-text retrieval over RGB-only and non-IR-aware settings. Ablations show that IR-aware captions improve task-conditioned infrared description, demonstrating the value of modality-specific supervision. FusionRS provides a scalable foundation for controlled RGB--infrared remote sensing vision--language learning.
\end{abstract}



\maketitle

\noindent\textbf{Code:} \href{https://github.com/frozy129/FusionRS}{github.com/frozy129/FusionRS}

\begin{table*}[t!]
\centering
\caption{
Comparison between our dataset and existing RGB--IR, infrared, and remote sensing vision-language datasets.
Existing RGB--IR datasets mainly focus on surveillance, driving, fusion, detection, or segmentation, while existing remote sensing vision-language datasets are mostly limited to RGB--text pairs.
In contrast, FusionRS provides large-scale RGB--infrared-style--text triplets for remote sensing, enabling controlled shared-embedding learning, task-conditioned caption adaptation, and a held-out IR-answerable VQA diagnostic.
}
\label{tab:dataset_comparison}
\resizebox{\textwidth}{!}{
\begin{tabular}{c>{\centering\arraybackslash}m{2.8cm}>{\centering\arraybackslash}m{4.8cm}>{\centering\arraybackslash}m{6.2cm}}
\toprule
\textbf{Dataset} & \textbf{Size} & \textbf{Source \& Domains} & \textbf{Coverage} \\
\midrule
LLVIP~\cite{jia2021llvip}
& 15K pairs
& Visible-infrared low-light surveillance scenes
& RGB--IR pairing, pedestrian detection, image fusion \\

M3FD~\cite{liu2022target}
& 4K pairs
& Multi-scenario visible-infrared street scenes
& RGB--IR fusion, object detection \\

MFNet~\cite{ha2017mfnet}
& 1.6K pairs
& RGB-thermal urban driving scenes
& RGB-T semantic segmentation \\[2pt]

FLIR Thermal ADAS~\cite{fliradas}
& \makecell{9,711 thermal \\ 9,233 RGB images}
& Thermal and visible driving scenes
& Object detection, autonomous driving \\

Infrared-LLaVA~\cite{jiang2024infrared}
& \makecell{118K synthetic IR \\ 500K IR--text pairs}
& COCO images translated with sRGB-TIR
& General IR VLM alignment, 12K instructions, real-IR benchmark \\

IRGPT~\cite{cao2025irgpt}
& \makecell{260K+ authentic \\ IR--text pairs}
& 63 real infrared datasets across multiple domains
& Infrared VLM pretraining, instruction tuning, nine-task benchmark \\

\midrule
VEDAI~\cite{razakarivony2016vehicle}
& 1.2K images
& Aerial vehicle images with multiple spectral bands
& Remote sensing vehicle detection \\

DroneVehicle~\cite{sun2022drone}
& 28K pairs
& Drone-based RGB-infrared vehicle images
& UAV traffic scenes, vehicle detection \\

RSICD~\cite{lu2017exploring}
& 10.9K images
& Remote sensing images from Google Earth and map platforms
& RGB--text captioning, image-text retrieval \\

RSITMD~\cite{yuan2021exploring}
& 4.7K images
& Remote sensing images from RSICD~\cite{lu2017exploring} and Google Earth
& Fine-grained RGB--text retrieval \\

RS5M~\cite{zhang2024rs5m}
& 5M pairs
& Large-scale remote sensing image-text pairs
& Remote sensing vision-language pretraining \\

\midrule
\textbf{FusionRS}
& \textbf{600K aligned records}
& \textbf{RGB remote sensing images translated to infrared-style views}
& \textbf{RGB--IR--text triplets, group-aware splits, IR-aware captions, retrieval and real-sensor transfer} \\
\bottomrule
\end{tabular}
}
\end{table*}

\section{Introduction}

Remote sensing imagery supports land-cover mapping, urban monitoring, disaster assessment, and agricultural observation. Vision-language models now enable open-vocabulary retrieval, caption generation, visual question answering, and instruction following beyond fixed taxonomies. Large-scale datasets such as RemoteCLIP~\cite{liu2024remoteclip}, GeoRSCLIP~\cite{zhang2024rs5m}, and SkyScript~\cite{wang2024skyscript} have driven progress in aligning RGB imagery with natural language~\cite{kuckreja2024geochat,bazi2024rs,wang2024ringmogpt}. Infrared sensors provide intensity, contrast, and thermal measurements that complement visible-spectrum imagery, but physical signals absent from RGB cannot be recovered by image translation alone. Scalable infrared-style data can nevertheless support controlled study of appearance shift and cross-view alignment before transfer is evaluated on sensor-captured infrared benchmarks. Yet paired RGB--infrared remote sensing data with language annotations remain scarce at scale.

Table~\ref{tab:dataset_comparison} positions FusionRS against existing RGB--infrared datasets and remote sensing vision-language corpora. Ground-level RGB--IR datasets (LLVIP, M3FD, FLIR) target surveillance and driving with authentic thermal pairs but remain small-scale and domain-specific, limiting their applicability to aerial Earth observation scenarios. Infrared vision-language efforts (Infrared-LLaVA, IRGPT) provide text supervision yet focus on ground-level scenes or lack RGB alignment. Remote sensing datasets (RSICD, RS5M) scale to millions of image--text pairs but omit infrared modalities entirely. FusionRS instead combines 600K aligned RGB--infrared-style pairs, full-corpus scene text, a curated IR-aware subset, group-aware splits, and separate sensor-captured transfer evaluation, bridging the gap between large-scale pretraining data and infrared-aware remote sensing applications.

Our benchmark trains CLIP-style encoders that embed RGB views, infrared-style views, and text in a unified space through trimodal contrastive objectives. We evaluate cross-view retrieval at scale across nested training subsets and controlled visual-view ablations, then measure transfer to sensor-captured RGB--thermal data from HIT-UAV and Caltech through zero-shot, frozen-probe, adapted, and synchronized-pair protocols. These protocols isolate representation quality from task-specific fine-tuning effects. A complementary generative evaluation tests Caption/VQA models on 499 infrared-only queries whose answers are observable in the infrared-style input, assessing task-conditioned caption control and zero-shot visual reasoning. Our main contributions are summarized as follows:

\begin{figure*}[t]
\centering
\includegraphics[width=0.98\textwidth]{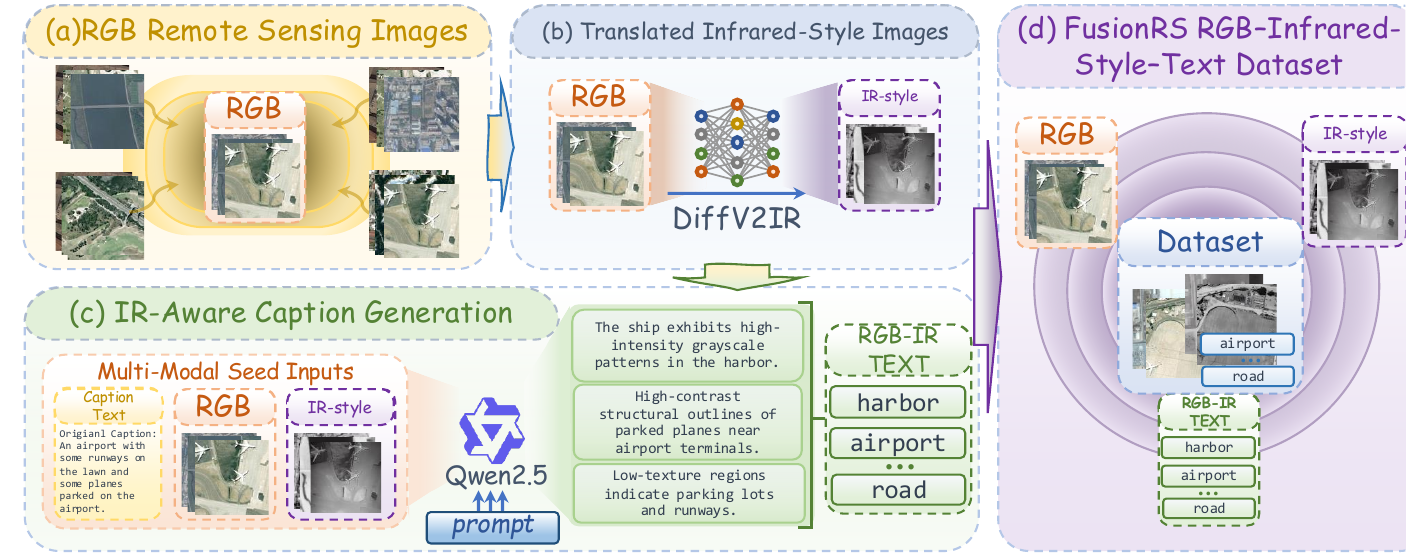}
\caption{
Overview of the proposed RGB--infrared-style--text dataset construction pipeline.
FusionRS is constructed by collecting large-scale RGB remote sensing images, translating them into infrared-style observations with an RGB-to-IR diffusion translator, and generating IR-aware textual descriptions from RGB images, IR images, and original scene-level text.
The resulting RGB images, synthetic infrared-style images, and source captions form the primary shared-embedding benchmark; IR-aware captions support a separately gated generative evaluation.
}
\Description{FusionRS construction pipeline from public RGB remote sensing images to translated infrared-style images and language annotations.}
\label{fig:dataset_pipeline}

\end{figure*}

\begin{itemize}
    \item To the best of our knowledge, we present the first large-scale RGB--infrared-style--text dataset for remote sensing, comprising 579,992 training triplets and 45,913 infrared-aware captions. The approach translates diverse RGB remote sensing imagery into aligned infrared-style views via diffusion-based synthesis, then generates textual descriptions that capture observable intensity, contrast, texture, and structural patterns while preserving scene semantics.

    \item We conduct extensive experiments that validate FusionRS strengthens cross-modal alignment: three full-scale training seeds achieve $95.35\pm0.04$\% paired-view Mean Recall. Sensor-captured HIT-UAV and Caltech evaluations cover zero-shot transfer, frozen probes, downstream detector adaptation, and synchronized RGB--thermal retrieval, revealing strong correspondence gains and task-dependent semantic transfer.

    \item We quantify both utility and representativeness limits. Direct RGB--infrared-style alignment lifts paired-view Mean Recall from 66.97\% to 95.31\%, whereas comparison with sensor-captured HIT-UAV shows similar mean brightness ($d=0.07$) but substantial shifts in other appearance statistics ($d=0.92$--$2.03$). FusionRS is therefore infrared-style pretraining data, not physical thermal ground truth.
\end{itemize}

\section{Related Work}

\paragraph{Vision-language pretraining.}
Contrastive vision-language pretraining aligns visual and textual representations through paired supervision. CLIP~\cite{radford2021learning} introduced large-scale contrastive learning on 400 million image--text pairs, demonstrating strong zero-shot transfer across downstream tasks. ALIGN~\cite{jia2021scaling} scaled to 1.8 billion pairs, while OpenCLIP~\cite{cherti2023reproducible} studied scaling laws across data volume and model capacity. These methods maximize matched-pair similarity in dual-encoder architectures through symmetric cross-entropy objectives. Generative models further extend this paradigm: BLIP~\cite{li2022blip} combines contrastive learning with captioning objectives, while BLIP-2~\cite{li2023blip} introduces query transformers to bridge frozen vision and language models. These strategies form the foundation for domain transfer, yet their application to RGB--infrared remote sensing remains limited by scarce paired cross-modal supervision.

\paragraph{Remote sensing vision-language learning.}
Vision-language models have adapted remote sensing to retrieval~\cite{lu2017exploring,yuan2021exploring}, visual question answering~\cite{lobry2020rsvqa}, and instruction following~\cite{kuckreja2024geochat}. Early datasets provided narrow supervision: RSICD~\cite{lu2017exploring} (10.9K pairs) and RSITMD~\cite{yuan2021exploring} (4.7K pairs). Large-scale pretraining---RS5M (5 million)~\cite{zhang2024rs5m}, SkyScript (2.6 million)~\cite{wang2024skyscript}, and RemoteCLIP~\cite{liu2024remoteclip}---improved zero-shot transfer by scaling contrastive objectives to millions of aerial observations. Multimodal LLMs (GeoChat~\cite{kuckreja2024geochat}, RS-LLaVA~\cite{bazi2024rs}, RingMoGPT~\cite{wang2024ringmogpt}) further enabled open-ended reasoning. Despite these advances, visible-only pretraining does not directly address infrared appearance and sensor domains. The absence of large-scale RGB--infrared--text triplets has constrained cross-modal retrieval and representation learning in Earth observation.

\paragraph{RGB-infrared multimodal learning.}
Paired RGB-infrared datasets support detection and segmentation in surveillance, driving, and aerial scenes (LLVIP~\cite{jia2021llvip}, FLIR~\cite{fliradas}, MFNet~\cite{ha2017mfnet}, M3FD~\cite{liu2022target}). Translation methods reduce reliance on costly thermal acquisition through edge guidance, self-supervised consistency, or style control~\cite{lee2023edge,sikdar2024ssl,ding2026synth}. IRGPT~\cite{cao2025irgpt} provides 260K authentic infrared--text pairs, whereas Infrared-LLaVA~\cite{jiang2024infrared} uses translated ground-level imagery. MonoIR-RS~\cite{han2026monoir} reuses the FusionRS source pool and split to isolate IR-only CLIP/VLM adaptation; FusionRS instead retains aligned RGB--infrared-style--text triplets and explicit cross-view objectives. The two settings separate single-modal IR adaptation from paired cross-modal learning.

\paragraph{Our positioning.}
FusionRS is the first large-scale RGB--infrared-style remote sensing dataset with aligned text. It translates 600K RGB images, retains source captions for the full contrastive benchmark, and adds IR-aware descriptions to a curated subset. Explicit image--text and cross-view objectives support paired retrieval, while zero-shot, frozen-probe, detector-adaptation, and synchronized-retrieval protocols test transfer to sensor-captured data.

\begin{figure*}[t]
\centering
\fontsize{7.0}{8.3}\selectfont
\setlength{\tabcolsep}{3pt}
\renewcommand{\arraystretch}{1.04}
\begin{tabular}{C{0.075\textwidth}C{0.145\textwidth}C{0.145\textwidth}C{0.53\textwidth}}
\toprule
\textbf{Source} & \textbf{RGB} & \textbf{IR-style} & \textbf{Original / IR-aware annotation} \\
\midrule
NWPU airport &
\includegraphics[width=\linewidth]{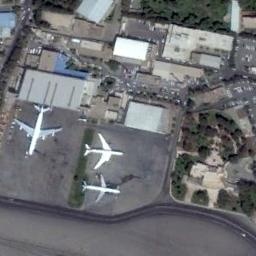} &
\includegraphics[width=\linewidth]{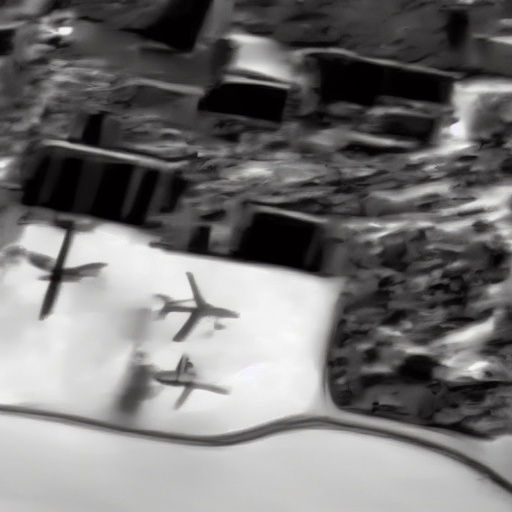} &
\textbf{Original:} Three planes are parked in an open area, with several buildings nearby. \newline
\textbf{IR-aware:} Three dark airplane silhouettes lie on a broad bright paved area, with blocky darker structures along the upper edge. \\
\midrule
RS5M courts &
\includegraphics[width=\linewidth]{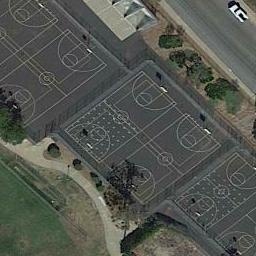} &
\includegraphics[width=\linewidth]{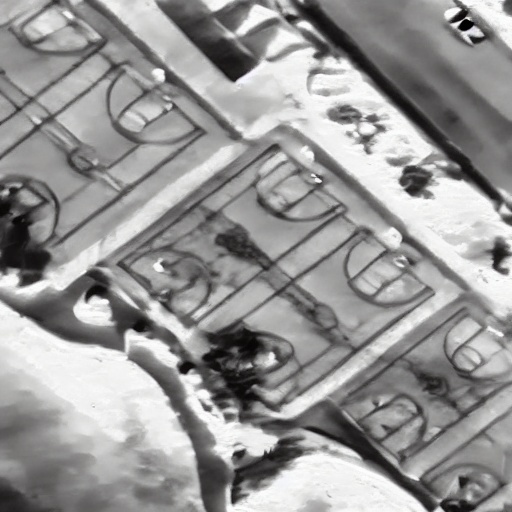} &
\textbf{Original:} Aerial image of basketball courts, tennis courts, and a football court. \newline
\textbf{IR-aware:} Several rectangular courts show bright line markings, dark interior surfaces, and repeated circular center features beside a linear road. \\
\midrule
RSITMD interchange &
\includegraphics[width=\linewidth]{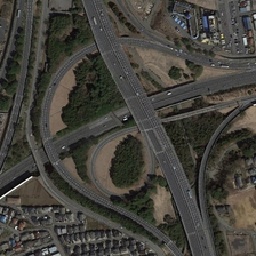} &
\includegraphics[width=\linewidth]{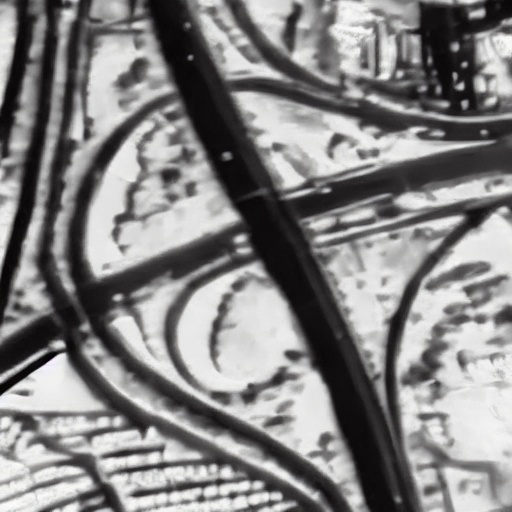} &
\textbf{Original:} Roads and viaducts form a cross across the saddle line. \newline
\textbf{IR-aware:} Dark and bright roadway bands intersect diagonally, with curved ramps enclosing irregular lighter ground areas. \\
\bottomrule
\end{tabular}
\caption{
FusionRS annotation examples. Each row shows an RGB source image, its translated infrared-style view, and both the original scene caption and the corresponding IR-aware annotation.
}
\Description{Three representative FusionRS records with RGB images, translated infrared-style images, original captions, and filtered infrared-aware captions.}
\label{fig:dataset_examples}
\end{figure*}

\section{Dataset Construction}
\label{sec:dataset_construction}

Table~\ref{tab:dataset_statistics} summarizes the dataset scale, source-balanced partitioning, and generative supervision. RS5M provides the largest share of the 600K aligned records, while SkyScript, NWPU-Captions, RSICD, and RSITMD contribute complementary scenes and caption styles. The group-aware assignment preserves the 580K/10K/10K train/validation/test split, and the curated subset provides 45,913 IR-aware captions and 45,912 matched records for generative training.

\begin{table}[t]
\centering
\caption{FusionRS source, split, and annotation summary. ``IR-aware'' counts curated infrared-aware captions; ``VLM'' counts records with matched source and IR-aware supervision.}
\label{tab:dataset_statistics}
\setlength{\tabcolsep}{3.5pt}
\scriptsize
\resizebox{\columnwidth}{!}{
\begin{tabular}{lcccc}
\toprule
\textbf{Source} & \textbf{Total} & \textbf{Train/Val/Test} & \textbf{IR-aware} & \textbf{VLM} \\
\midrule
RS5M~\cite{zhang2024rs5m} & 488,033 & 472,682/7,670/7,681 & 38,411 & 38,410 \\
SkyScript~\cite{wang2024skyscript} & 65,266 & 64,150/558/558 & 4,593 & 4,593 \\
NWPU-Captions~\cite{cheng2022nwpucaptions} & 31,186 & 30,094/556/536 & 2,014 & 2,014 \\
RSICD~\cite{lu2017exploring} & 10,824 & 9,383/716/725 & 642 & 642 \\
RSITMD~\cite{yuan2021exploring} & 4,691 & 3,691/500/500 & 253 & 253 \\
\midrule
\textbf{Total} & \textbf{600,000} & \textbf{580,000/10,000/10,000} & \textbf{45,913} & \textbf{45,912} \\
\bottomrule
\end{tabular}
}
\end{table}

FusionRS contains 600,000 aligned records constructed from diverse public RGB remote sensing datasets. Each record pairs an RGB image with a translated infrared-style view. Of the 600,000 records, 599,992 include usable source-text supervision; the training partition contains 580,000 records, of which 579,992 carry valid text, forming the aligned triplets used for CLIP training. Figure~\ref{fig:dataset_pipeline} presents the construction pipeline, while Table~\ref{tab:dataset_statistics} gives the compact source, split, and annotation summary. Complete counts are provided in Appendix Table~\ref{tab:app_source_iraware_statistics}.

\subsection{Data Sources and Splits}
\label{subsec:data_sources}

FusionRS draws from five public image--text datasets: RS5M~\cite{zhang2024rs5m}, SkyScript~\cite{wang2024skyscript}, NWPU-Captions~\cite{cheng2022nwpucaptions}, RSICD~\cite{lu2017exploring}, and RSITMD~\cite{yuan2021exploring}. Together, they cover airports, roads, residential areas, farmland, industrial regions, water bodies, and other common Earth observation scenes. RS5M supplies most records, while the other sources broaden scene and caption diversity. The public release is index-only, providing source identifiers, group-aware split assignments, reconstruction scripts, configurations, checksums, a Datasheet-style card, and Croissant metadata without redistributing upstream RGB or translated infrared-style images. Generated annotations are released only for sources with confirmed rights; a machine-readable table records source-level clearance, and users obtain upstream data through official channels and materialize indexed views locally under the original terms.

We partition 600,000 samples into 580,000 training, 10,000 validation, and 10,000 test records. A joint group-aware procedure links similar RGB images, infrared-style images, and normalized captions before assigning each connected group to a single partition. The assignment preserves the source distribution and places at least 500 records from every source in each held-out split. In-domain evaluation uses one representative per held-out group, producing 9,538 validation and 9,553 test triplets.

\begin{figure*}[t]
\centering
\includegraphics[width=0.98\textwidth]{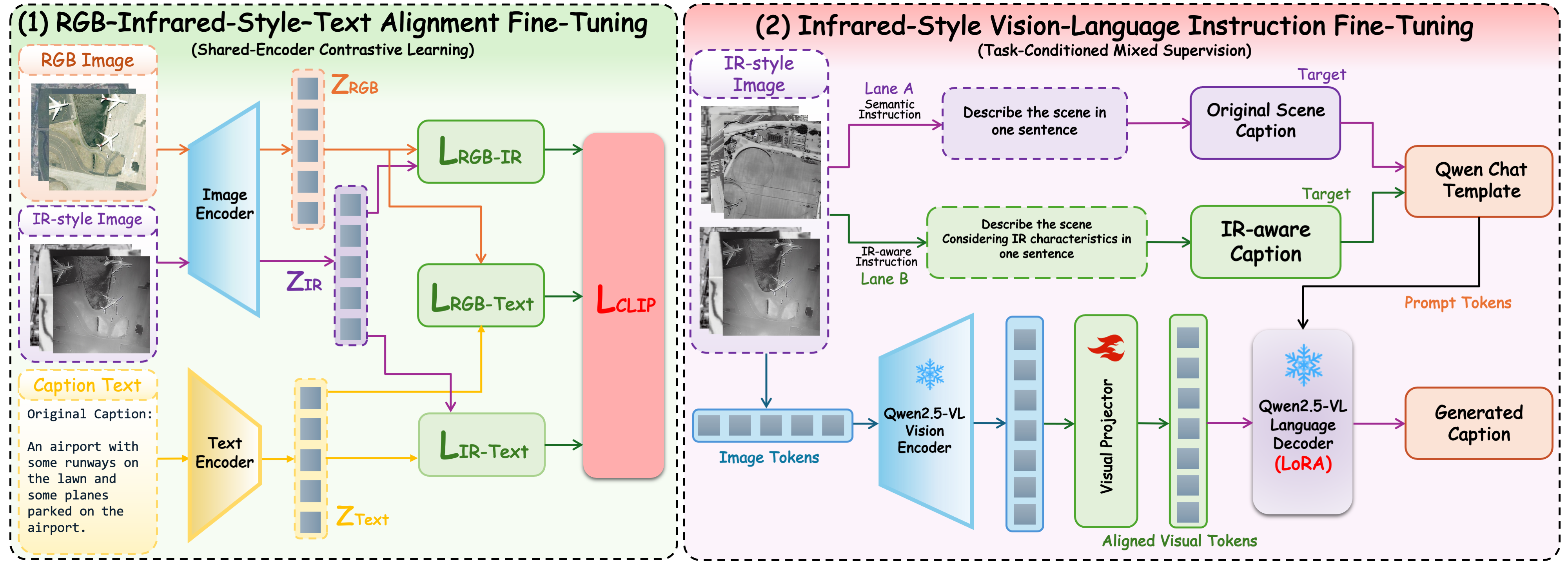}
\caption{
FusionRS evaluation pipeline. Stage 1 uses a shared CLIP image encoder for RGB and infrared-style views with dual-text or trimodal objectives. Stage 2 uses infrared-style images, task-specific prompts, and assistant-only caption supervision; VQA is evaluated only as a held-out zero-shot diagnostic.
}
\Description{Two-stage training pipeline for shared-embedding CLIP alignment and single-image infrared-style VLM adaptation.}
\label{fig:training_pipeline}
\end{figure*}

\subsection{RGB-to-IR Image Translation}
\label{subsec:rgb_to_ir}

We convert RGB remote sensing images into infrared-style observations using DiffV2IR~\cite{ran2025diffv2ir}, a diffusion-based visible-to-infrared translation model. The published Phase-2 checkpoint produces a spatially aligned infrared-style view for each RGB image. Generated infrared-style images are saved as \texttt{.jpg} files in PIL RGB mode rather than forced single-channel grayscale. Their stored dimensions follow the materialized translation outputs; model-specific processors apply the required resize and crop. Visually, the images follow a grayscale infrared style, but the three channels are not forced to be identical. This format allows direct use by standard CLIP-style encoders and generative VLMs without modifying their input interfaces. Through this translation process, each RGB image is paired with an infrared-style counterpart. The aligned views retain scene layout while introducing intensity-dominant appearance, reduced color texture, stronger structural contrast, and infrared-style object boundaries.

\subsection{IR-aware Text Annotation and Quality Control}
\label{subsec:ir_caption}

FusionRS contains 599,992 source captions or scene-level descriptions that provide semantic supervision for image--text alignment. Because original RGB captions seldom describe grayscale intensity, contrast, texture, or structural outlines, we generate complementary IR-aware captions using Qwen2.5-VL-72B-Instruct~\cite{bai2025qwen25vltechnicalreport} through the OpenRouter API. Given the translated infrared-style image and original text, the model produces a caption that preserves scene semantics while describing observable intensity, contrast, bright and dark regions, object outlines, and texture. The release freezes all 45,913 accepted caption strings together with the exact model-route identifier, prompt, decoding settings, and May 2026 access period. Public caption files remain subject to the source-rights gate described above. Because the hosted route exposed neither an immutable model snapshot nor a pinned underlying provider, we do not claim byte-identical text regeneration. Figure~\ref{fig:dataset_examples} shows representative RGB--infrared-style pairs with their original and IR-aware annotations. We retain captions that satisfy semantic, infrared-cue, format, and content checks. Of the 45,913 curated IR-aware captions, 45,912 have matched source captions for generative supervision, yielding 91,824 task-conditioned instruction--response records. The 50K--580K CLIP benchmark uses source captions for RGB--text and infrared-style--text learning, while the IR-aware captions support task-conditioned caption adaptation. The separately constructed VQA benchmark is never used for adapter training and is retained only for zero-shot diagnosis in Appendix~\ref{app:vlm_results}.

Source captions are normalized and validated before CLIP training, while IR-aware captions pass format, semantic-consistency, IR-cue, and language-quality checks. The two supervision streams remain separate: source captions define the primary contrastive benchmark, while IR-aware captions support generative training. Quantitative image and caption audits are reported in Section~\ref{subsec:quality_audit_results}; Appendix Tables~\ref{tab:app_qualitative_examples_a}--\ref{tab:app_qualitative_examples_h} provide additional representative records.

\section{Model Training and Experimental Evaluation}
\label{sec:training_protocol}

FusionRS uses a two-stage framework connecting cross-modal representation learning with infrared-grounded generation. Stage~1 continues contrastive pretraining from OpenAI CLIP using aligned RGB--infrared-style--text triplets. A shared image encoder maps RGB and infrared-style views into one embedding space, while the text encoder represents their common scene description. Three coupled objectives align RGB with text, infrared-style imagery with text, and the paired visual views with each other, promoting semantic consistency and modality-invariant representations. Shared visual parameters allow either modality to be processed independently at inference; aligned pairs are required only during training. Stage~2 adapts Qwen2.5-VL-7B-Instruct~\cite{bai2025qwen25vltechnicalreport} with assistant-only label masking and two task-conditioned targets: source-scene and IR-aware captions. Evaluation is strictly IR-only: captioning measures the supervised target, while VQA remains a zero-shot diagnostic because no VQA instruction--answer pairs are used for adaptation.

The main controlled study uses OpenAI CLIP ViT-B/32~\cite{radford2021learning}; Appendix Table~\ref{tab:app_capacity} provides a matched ViT-B/32 versus ViT-L/14 capacity analysis. RemoteCLIP~\cite{liu2024remoteclip} and GeoRSCLIP~\cite{zhang2024rs5m} are evaluated as zero-shot remote-sensing initializations. Figure~\ref{fig:training_pipeline} shows the framework, while Tables~\ref{tab:clip_main_results} and~\ref{tab:clip_ablation_scaling} separate the overall comparison from controlled data-scale, objective, and visual-view analyses. Retrieval uses 9,538 validation and 9,553 test representatives fixed by the group-aware manifest with normalized-caption multi-positive matching. The released scorer verifies representative identities and computes bidirectional R@1/5/10 and the Text, Pair, and Overall means. Sensor-captured transfer follows official splits, validation-only model selection, and a single test evaluation; table captions specify the seed protocols.

\begin{table}[t]
\centering
\caption{Retrieval overview with off-the-shelf reference models. Text Mean averages bidirectional IR-style--text R@1/5/10, Pair Mean averages bidirectional RGB--IR-style R@1/5/10, and Overall averages the two. OpenAI CLIP, RemoteCLIP, and GeoRSCLIP are zero-shot references rather than matched-budget training baselines; FusionRS reports mean $\pm$ SD over three seeds.}
\label{tab:clip_main_results}
\setlength{\tabcolsep}{4pt}
\small
\resizebox{\columnwidth}{!}{
\begin{tabular}{lccc}
\toprule
\textbf{Method} & \textbf{Text Mean} & \textbf{Pair Mean} & \textbf{Overall} \\
\midrule
OpenAI CLIP~\cite{radford2021learning} & 3.34 & 9.52 & 6.43 \\
RemoteCLIP~\cite{liu2024remoteclip} & 0.92 & 5.06 & 2.99 \\
GeoRSCLIP~\cite{zhang2024rs5m} & 5.89 & 12.96 & 9.43 \\
\midrule
\textbf{FusionRS} & $\mathbf{40.23\pm0.20}$ & $\mathbf{95.35\pm0.04}$ & $\mathbf{67.79\pm0.12}$ \\
\bottomrule
\end{tabular}
}
\end{table}

\subsection{RGB--Infrared-Style--Text CLIP Training and Retrieval}
\label{subsec:clip_training}

Training uses 579,992 aligned triplets. Each sample retains its RGB image, translated infrared-style view, and source caption, so all three positive relations are available within the same optimization step. The nested 50K, 100K, and 300K subsets preserve the source proportions of the full training set and share the same validation and test groups. Fine-tuning starts from the pretrained OpenAI CLIP ViT-B/32 weights. RGB and infrared-style images pass independently through the same visual encoder, while the caption is encoded once by the text tower. Each triplet therefore produces RGB--text, infrared-style--text, and RGB--infrared similarity matrices without modality-specific adapters. The shared encoder permits either modality to be embedded independently at inference.

Let $\mathcal{B}=\{(r_i,v_i,c_i)\}_{i=1}^{N}$ be a mini-batch of $N$ triplets, where $r_i$, $v_i$, and $c_i$ denote the RGB image, infrared-style image, and caption of sample $i$. The shared CLIP image encoder $f_\theta$ and text encoder $g_\phi$ include their respective projection layers. Denoting L2 normalization by $\mathcal{N}(\mathbf{x})=\mathbf{x}/\lVert\mathbf{x}\rVert_2$, the embeddings are
\begin{equation}
\mathbf{z}_i^r=\mathcal{N}(f_\theta(r_i)),\quad
\mathbf{z}_i^v=\mathcal{N}(f_\theta(v_i)),\quad
\mathbf{z}_i^c=\mathcal{N}(g_\phi(c_i)).
\end{equation}
The same $f_\theta$ processes both visual modalities. For modalities $a,b\in\{r,v,c\}$, the learnable CLIP logit scale $\alpha$ produces the pairwise similarity
\begin{equation}
S_{a,b}^{ij}
=
\exp(\alpha)\,(\mathbf{z}_i^a)^\top\mathbf{z}_j^b,
\qquad i,j\in\{1,\ldots,N\}.
\end{equation}
Diagonal entries are matched pairs, while non-diagonal batch elements serve as negatives. Both retrieval directions are optimized through the symmetric loss
\begin{equation}
\mathcal{L}_{a,b}
=-\frac{1}{2N}\sum_{i=1}^{N}\left[
\log\frac{e^{S_{a,b}^{ii}}}{\sum_{j=1}^{N}e^{S_{a,b}^{ij}}}
+\log\frac{e^{S_{a,b}^{ii}}}{\sum_{j=1}^{N}e^{S_{a,b}^{ji}}}
\right].
\end{equation}
The two directions exchange query and target roles. With 128 triplets, every similarity matrix contains 128 positives and 16,256 in-batch negatives, providing dense supervision without an external queue.

\begin{table}[t]
\centering
\caption{Matched CLIP supervision, visual-view, and data-scale ablations. All rows use OpenAI CLIP ViT-B/32 initialization, the same held-out groups, and seed 42; the full three-seed result is reported in Table~\ref{tab:clip_main_results}. Text Mean and Pair Mean retain language retrieval and paired-view retrieval as separate objectives.}
\label{tab:clip_ablation_scaling}
\setlength{\tabcolsep}{2pt}
\small
\begin{tabular}{@{}p{0.515\columnwidth}C{0.11\columnwidth}C{0.155\columnwidth}C{0.155\columnwidth}@{}}
\toprule
\textbf{Training setting} & \textbf{Scale} & \mbox{\textbf{Text Mean}} & \mbox{\textbf{Pair Mean}} \\
\midrule
\mbox{RGB-only training} & \multirow{4}{*}{580K} & 10.05 & 18.72 \\
\mbox{RGB duplicate, trimodal loss} & & 10.03 & 18.85 \\
\mbox{Grayscale view, trimodal loss} & & 13.81 & 28.69 \\
\mbox{Translated view, dual-text loss} & & \textbf{40.73} & 66.97 \\
\midrule
\multirow{4}{=}{\mbox{Translated view, trimodal loss}} & 50K & 22.87 & 85.92 \\
& 100K & 27.21 & 89.51 \\
& 300K & 35.31 & 93.54 \\
& 580K & 40.00 & \textbf{95.31} \\
\bottomrule
\end{tabular}
\vspace{-4pt}
\end{table}

Applying this loss to RGB--text, infrared-style--text, and RGB--infrared-style pairs gives
\begin{equation}
\mathcal{L}_{\mathrm{FusionRS}}(\lambda)
=
\frac{
\mathcal{L}_{r,c}
+
\mathcal{L}_{v,c}
+
\lambda\mathcal{L}_{r,v}
}{2+\lambda},
\qquad \lambda=1.
\end{equation}
The two image--text terms preserve scene-level language alignment, while $\mathcal{L}_{r,v}$ supplies instance-level cross-view correspondence. Equal weighting ($\lambda=1$) defines the full trimodal model. Setting $\lambda=0$ yields the \emph{dual-text} control while retaining the same triplets, visual views, and optimization exposure. Further controls replace $v_i$ with deterministic grayscale RGB or an exact RGB duplicate, isolating infrared-style appearance from generic two-view training. The three objectives are combined in one backward pass, jointly updating both encoders, projection layers, and the logit scale with AdamW. Full-scale training uses batches of 128 triplets (256 image views and 128 captions), a learning rate of $5\times10^{-6}$, weight decay $0.1$, 500 warm-up steps followed by cosine decay, mixed-precision training, and three epochs (13,593 optimizer steps). Validation loss is computed after each epoch. Unless otherwise stated, ablations use seed 42; the full trimodal result is repeated with seeds 42, 3407, and 2026.

Table~\ref{tab:clip_main_results} evaluates 9,553 unique test representatives. The off-the-shelf rows quantify zero-shot behavior on the FusionRS protocol and are not used as matched-training evidence. Across three continued-pretraining seeds, FusionRS reaches $40.23\pm0.20$ Text Mean and $95.35\pm0.04$ Pair Mean. The corresponding controlled evidence is Table~\ref{tab:clip_ablation_scaling}, where initialization, held-out groups, and training exposure are matched. Table~\ref{tab:clip_ablation_scaling} isolates supervision, view, and scale. At 580K, alignment raises Pair Mean from 66.97 to 95.31. Scaling translated data from 50K to 580K raises Text Mean from 22.87 to 40.00 and Pair Mean from 85.92 to 95.31. Translated views outperform all controls. Similar RGB-only and duplicate results exclude extra-input effects; smaller grayscale gains indicate infrared-style appearance. Gains therefore derive from infrared-style appearance and cross-view supervision. Full Recall@1/5/10, per-source results, and seed robustness are in Appendix Tables~\ref{tab:app_full_retrieval}--\ref{tab:app_per_source_retrieval}.

\FloatBarrier

\begin{table*}[t]
\centering
\caption{Task-conditioned caption ablation on 499 held-out infrared-style images. Semantic and IR-cue R-L denote task-specific ROUGE-L. Semantic Cue is IR-cue leakage under the semantic prompt; IR Cue is cue coverage under the IR-aware prompt; Cue Gap is their difference. Unsupported Physical is a rule-based lexical flag rate on IR-aware outputs, not a human factual-grounding judgment. Prompt Collapse is the fraction of paired prompts producing identical outputs. All rates are percentages.}
\label{tab:vlm_main_results}
\setlength{\tabcolsep}{3.5pt}
\small
\begin{tabular*}{\textwidth}{@{\extracolsep{\fill}}lccccccc@{}}
\toprule
\textbf{Setting} & \makecell{\textbf{Semantic}\\\textbf{R-L}} & \makecell{\textbf{IR-cue}\\\textbf{R-L}} & \makecell{\textbf{Semantic}\\\textbf{Cue} $\downarrow$} & \makecell{\textbf{IR}\\\textbf{Cue} $\uparrow$} & \makecell{\textbf{Cue}\\\textbf{Gap} $\uparrow$} & \makecell{\textbf{Unsupported}\\\textbf{Physical} $\downarrow$} & \makecell{\textbf{Prompt}\\\textbf{Collapse} $\downarrow$} \\
\midrule
Qwen2.5-VL-7B (base) & \textbf{20.45} & \textbf{19.74} & 99.60 & 91.18 & $-8.42$ & 16.63 & \textbf{0.00} \\
Original-only & 18.84 & 14.27 & 5.41 & 6.61 & 1.20 & \textbf{0.00} & 38.88 \\
IR-aware-only & 15.99 & 16.21 & 99.00 & \textbf{100.00} & 1.00 & \textbf{0.00} & 0.40 \\
\textbf{Mixed} & 19.31 & 16.05 & \textbf{4.41} & \textbf{100.00} & \textbf{95.59} & \textbf{0.00} & \textbf{0.00} \\
\bottomrule
\end{tabular*}
\end{table*}

\begin{figure*}[t!]
\centering
\includegraphics[width=\textwidth]{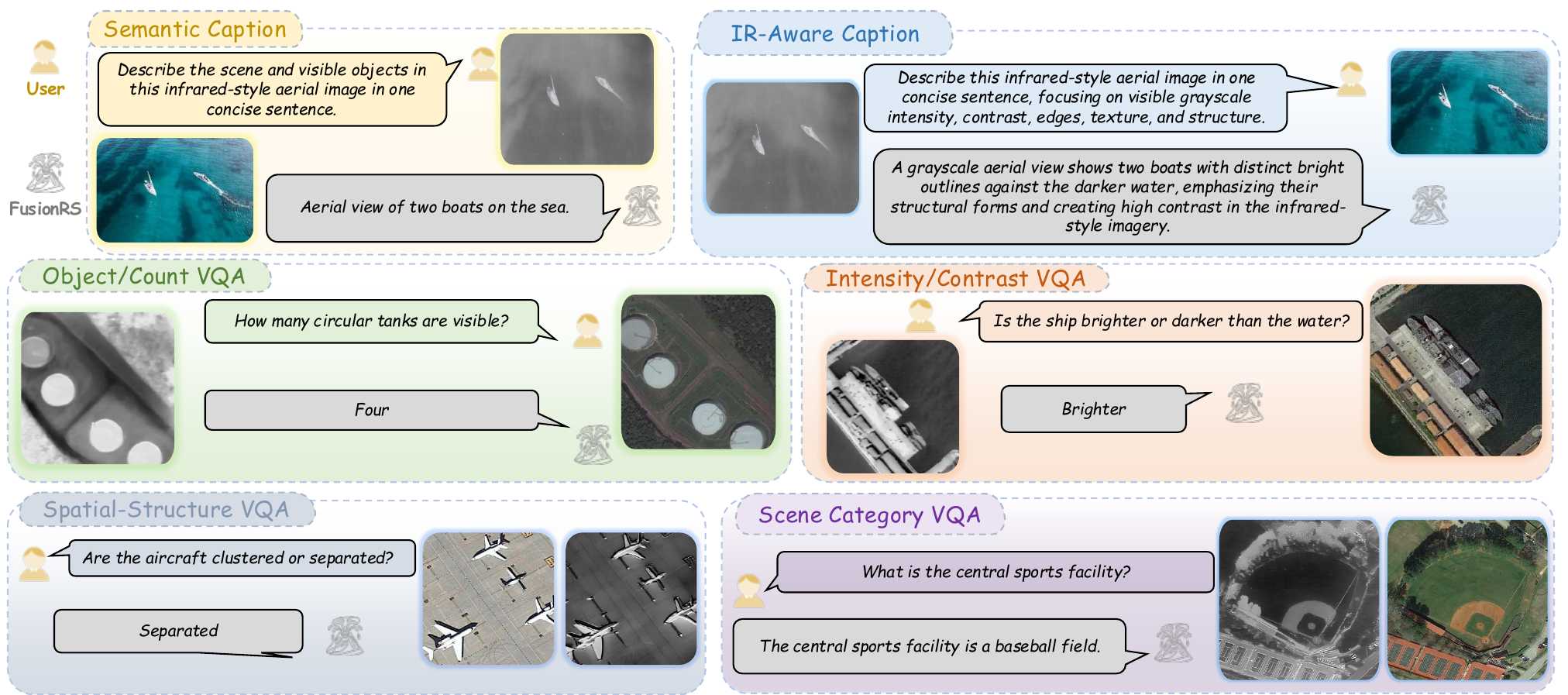}
\caption{\textbf{Task-conditioned IR-only Caption/VQA examples from the held-out set.} Each query provides a single infrared-style image and a task prompt; RGB views and references are unavailable at inference. The upper panels contrast semantic and IR-aware captioning, while the lower panels illustrate four IR-answerable VQA categories: object counting, intensity/contrast, spatial arrangement, and scene category. VQA is retained as a zero-shot diagnostic without VQA instruction tuning.}
\Description{Examples of semantic captioning, infrared-aware captioning, and four infrared-answerable visual question answering categories using infrared-style remote sensing images.}
\label{fig:task_conditioned_examples}
\end{figure*}

\subsection{Task-Conditioned VLM Adaptation and IR-Only Evaluation}
\label{subsec:vlm_evaluation}

We adapt Qwen2.5-VL-7B with 91,824 caption records from 45,912 training identities. Each infrared-style image yields two instruction--response records: a semantic prompt with its source-scene caption and an IR-aware prompt with the curated intensity-and-structure description. The \emph{Original-only} and \emph{IR-aware-only} controls retain one target, whereas \emph{Mixed} trains both using the same image encoder, projector, and LoRA-adapted decoder. At evaluation time, every model receives only an infrared-style image and one of the two prompts; RGB is available only to reference annotators. Figure~\ref{fig:task_conditioned_examples} shows the two caption conditions and four IR-answerable VQA categories used by the IR-only evaluation interface.

\begin{table*}[t]
\centering
\caption{\textbf{FusionRS image and caption quality audit.} Image statistics use a source-balanced sample of 200 training triplets. Caption retention is measured over the traceable accepted training pool; caption statistics are rule-based lexical diagnostics rather than human factual-grounding judgments.}
\label{tab:quality_audit}
\setlength{\tabcolsep}{7pt}
\small
\begin{tabular*}{\textwidth}{@{\extracolsep{\fill}}llcl@{}}
\toprule
\textbf{Component} & \textbf{Audit measure} & \textbf{Result} & \textbf{Interpretation} \\
\midrule
\multirow{5}{*}{Translated image} & Readable and spatially aligned & 200/200 (100\%) & No unreadable or visibly mismatched audited pair \\
& RGB--IR-style gradient correlation & 0.281 & Partial gradient-structure preservation \\
& RGB--IR-style edge-overlap F1 & 0.317 & Partial boundary preservation, not pixel identity \\
& Brightness gap vs.\ HIT-UAV & $d=0.07$ & Similar mean grayscale brightness ($n=200$ each) \\
& Other appearance gaps vs.\ HIT-UAV & $d=0.92$--$2.03$ & Contrast, range, entropy, edge, and sharpness shift \\
\midrule
\multirow{5}{*}{IR-aware caption} & Strict-manifest retention & 45,913/48,616 (94.44\%) & 2,703 rule-flagged records removed \\
& Mean caption length & 30.30 words & Detailed rather than label-only supervision \\
& IR-cue lexical coverage & 100\% & Predefined IR-style cue detected \\
& Source-token retention & 72.81\% & Lexical overlap with source caption; not semantic verification \\
& Class-name retention & 51.50\% & Explicit source class term retained \\
\bottomrule
\end{tabular*}
\vspace{-4pt}
\end{table*}

Table~\ref{tab:vlm_main_results} shows that ROUGE-L alone does not establish task control. The unadapted backbone obtains the highest ROUGE-L but inserts IR-specific language into 99.60\% of semantic responses and has a negative cue gap, failing the requested distinction. Original-only suppresses IR language in semantic responses but covers IR cues in only 6.61\% of IR-aware responses and produces identical outputs for 38.88\% of paired prompts. IR-aware-only instead inserts cues under both prompts, yielding a 1.00-point cue gap. Mixed is the only setting satisfying both conditions: semantic cue leakage falls to 4.41\%, IR-cue coverage reaches 100\%, and the cue gap reaches 95.59 points without prompt collapse. Under the same rule-based diagnostic, its unsupported-physical-language flag rate is 0\%, compared with 16.63\% for the base model. These lexical diagnostics support prompt-conditioned output separation, not factual or physical correctness. Mixed also retains the strongest semantic ROUGE-L among adapted variants and near-best IR-cue ROUGE-L. These results support the intended claim: mixed supervision enables one VLM to switch between conventional and IR-aware descriptions from the same input, but does not establish general VQA improvement. Since the adapters receive caption-only supervision, VQA remains a zero-shot diagnostic. Complete public-VLM Caption/VQA results, task-wise metrics, source-wise breakdowns, and annotation gates appear in Appendix~\ref{app:vlm_results}.

\begin{table}[H]
\centering
\caption{\textbf{Sensor-captured transfer.} HIT Det denotes mAP$_{50:95}$, HIT ZS zero-shot object-presence mAP, HIT LP frozen linear-probe mAP, and Caltech Pair bidirectional RGB--thermal Mean Recall. FusionRS controls use 580K records. Baseline HIT ZS, HIT LP, and Caltech Pair results use one seed; FusionRS ablations report mean $\pm$ SD when available.}
\label{tab:main_transfer}
\setlength{\tabcolsep}{3.5pt}
\small
\resizebox{\columnwidth}{!}{
\begin{tabular}{lcccc}
\toprule
\textbf{Initialization / training} & \textbf{HIT Det} & \textbf{HIT ZS} & \textbf{HIT LP} & \textbf{Caltech Pair} \\
\midrule
OpenAI CLIP~\cite{radford2021learning} & $35.81\pm1.40$ & 36.47 & 73.69 & 22.88 \\
RemoteCLIP~\cite{liu2024remoteclip} & $38.01\pm0.86$ & 38.71 & 79.22 & 15.29 \\
GeoRSCLIP~\cite{zhang2024rs5m} & $\mathbf{38.76\pm0.59}$ & 34.36 & 79.54 & 24.79 \\
\midrule
Translated, dual-text & $37.19\pm1.16$ & 39.88 & 76.90 & 43.43 \\
Grayscale, trimodal & $36.80\pm0.32$ & 39.24 & \textbf{81.85} & 35.36 \\
RGB duplicate, trimodal & $37.65\pm0.58$ & 37.05 & 80.04 & 33.09 \\
\textbf{FusionRS, trimodal} & $37.83\pm1.26$ & $\mathbf{40.70\pm0.46}$ & $79.09\pm2.20$ & \textbf{68.04} \\
\bottomrule
\end{tabular}
}
\vspace{-4pt}
\end{table}

\subsection{Image and Caption Quality Audit}
\label{subsec:quality_audit_results}

Table~\ref{tab:quality_audit} summarizes the dataset-quality evidence. A source-balanced audit of 200 training triplets finds every translated view readable and spatially aligned with its RGB source; gradient correlation and edge overlap indicate partial structural preservation. A separate random 200-vs.-200 comparison with sensor-captured HIT-UAV finds similar brightness ($d=0.07$) but substantial differences in contrast, dynamic range, entropy, edge strength, and Laplacian variance ($d=0.92$--$2.03$), quantifying an appearance gap rather than thermal representativeness. Caption filtering retains 45,913 of 48,616 traceable candidates; accepted captions contain predefined IR-style cue terms and retain 72.81\% of source tokens under lexical rules. These checks do not establish semantic correctness, temperature, emissivity, or material fidelity and do not replace blinded human evaluation.

\subsection{Sensor-Captured Transfer and Model Scaling}
\label{subsec:real_transfer}
We train OpenAI CLIP ViT-B/32 on nested 50K, 100K, 300K, and 580K subsets with fixed validation/test data and evaluate transfer on sensor-captured HIT-UAV~\cite{suo2023hituav} and Caltech Aerial RGB--thermal~\cite{lee2024caltechrgbt}. Table~\ref{tab:main_transfer} reports that translated trimodal training improves HIT-UAV detection from 35.81 to 37.83 mAP$_{50:95}$, zero-shot object-presence mAP from 36.47 to 40.70, linear-probe mAP from 73.69 to 79.09, and Caltech paired Mean Recall from 22.88 to 68.04 relative to OpenAI CLIP. These gains are control-relative rather than uniform state-of-the-art improvements: GeoRSCLIP remains stronger on HIT-UAV detection, while the grayscale trimodal control is stronger under the frozen linear probe. Because the frozen probe isolates last-layer separability, its result suggests that infrared-style texture may complicate linear boundaries without adding class-discriminative structure, whereas detection benefits from correspondence-driven full-backbone transfer. The largest FusionRS gain appears in synchronized RGB--thermal correspondence. Appendix Tables~\ref{tab:app_caltech_pair} and~\ref{tab:app_transfer_boundary} provide the full Caltech retrieval breakdown and additional adapted tasks.

\section{Scope and Ethical Considerations}
\label{sec:limitations}
FusionRS is designed for infrared-style representation learning, cross-view retrieval, and IR-grounded language evaluation in remote sensing research and development, with sensor-captured benchmarks providing complementary transfer measurements. Future extensions can incorporate additional translation models, authentic infrared sources, and broader human-reviewed generative tasks. Responsible use follows the licenses and access conditions of the five source datasets. As detailed in Section~\ref{subsec:data_sources}, the index-only release provides reconstruction manifests without redistributing upstream or translated images. Individual identification, persistent person or vehicle tracking, operational target localization or selection, autonomous surveillance, weapon-related decision making, and other safety-critical uses fall outside its scope. Requiring users to obtain source data and accept its terms allows institutional ethics and data-use review before high-stakes adoption.

\section{Conclusion}
\label{sec:conclusion}
FusionRS provides 600K aligned RGB--infrared-style records, 599,992 source-caption triplets, and 45,913 IR-aware captions. Retrieval improves monotonically with scale, and three trimodal seeds reach $95.35\pm0.04$\% paired-view Mean Recall. Direct RGB--infrared-style alignment yields the largest internal gain. Sensor-captured improvements are strongest on Caltech paired retrieval, while HIT-UAV detection and limited-label adaptation remain task-dependent. The dataset, controlled benchmark, and IR-only protocol support scalable RGB--infrared remote sensing vision--language learning.
\bibliographystyle{ACM-Reference-Format}
\bibliography{references}
\clearpage
\nobalance
\appendix
\raggedbottom

\section{Dataset Details}
\label{app:dataset_details}

This appendix provides additional construction details, complete ablation results, and qualitative examples for FusionRS.
Within FusionRS, ``IR'' denotes translated infrared-style remote sensing images; external benchmarks are explicitly identified as sensor-captured infrared or thermal data.

\subsection{Source Statistics}
\label{app:source_statistics}

\begin{table}[H]
\centering
\caption{\textbf{Complete source-wise split and annotation statistics.} The group-aware split assigns related RGB images, infrared-style images, and normalized captions to one partition. ``IR-aware'' counts curated captions, and ``VLM'' counts identities with matched source and IR-aware supervision.}
\label{tab:app_source_iraware_statistics}
\fontsize{6.8}{8.1}\selectfont
\setlength{\tabcolsep}{1.7pt}
\renewcommand{\arraystretch}{1.08}
\begin{tabular*}{\columnwidth}{@{\extracolsep{\fill}}lrrrrrr@{}}
\toprule
\textbf{Source} & \textbf{Total} & \textbf{Train} & \textbf{Val} & \textbf{Test} & \textbf{IR-aware} & \textbf{VLM} \\
\midrule
RS5M & 488,033 & 472,682 & 7,670 & 7,681 & 38,411 & 38,410 \\
SkyScript & 65,266 & 64,150 & 558 & 558 & 4,593 & 4,593 \\
NWPU-Captions & 31,186 & 30,094 & 556 & 536 & 2,014 & 2,014 \\
RSICD & 10,824 & 9,383 & 716 & 725 & 642 & 642 \\
RSITMD & 4,691 & 3,691 & 500 & 500 & 253 & 253 \\
\midrule
\textbf{Total} & \textbf{600,000} & \textbf{580,000} & \textbf{10,000} & \textbf{10,000} & \textbf{45,913} & \textbf{45,912} \\
\bottomrule
\end{tabular*}
\end{table}

Table~\ref{tab:app_source_iraware_statistics} makes the source composition and
annotation coverage explicit. RS5M supplies most records, whereas the remaining
sources broaden scene and caption diversity. The group-aware assignment preserves
the 580K/10K/10K split and yields 9,538 validation and 9,553 test representatives
after related records are collapsed for evaluation. IR-aware annotations cover
all five sources rather than only the dominant collection, and 45,912 of the
45,913 accepted captions map to valid VLM training identities. The table thus
connects the headline dataset size to the exact split, annotation, and
generative-training denominators used by the later experiments.

\FloatBarrier
\subsection{Release and Annotation Summary}
\label{app:release_annotation}

FusionRS follows source-specific terms and uses an index-only release. It
distributes source identifiers, group-aware splits, rights-cleared generated
annotations, reconstruction metadata and scripts, evaluation configurations,
and checkpoint hashes, but no upstream RGB or translated infrared-style image bytes. Users
obtain source images through official routes and materialize indexed views
locally under the applicable terms; translated derivatives are not
redistributed where derivative rights are unconfirmed, and source-caption text
and annotations conditioned on uncleared source material follow the same
source-specific treatment. A Datasheet-style card, Croissant
metadata, the source-rights table, schema, and fixed split hashes document
the release. Source captions provide semantic contrastive supervision, while
IR-aware captions preserve scene content and emphasize observable
infrared-style cues. Using Qwen2.5-VL-72B-Instruct, we curate 45,913 IR-aware
captions, match 45,912 of them to valid training identities, and construct
91,824 task-conditioned generative training records. The accepted caption
strings, exact OpenRouter route, prompt, decoding settings, and May 2026 access
period are frozen in the internal release candidate; public caption files remain
source-rights gated. Exact text regeneration is not claimed because the hosted
route did not expose an immutable snapshot or pinned provider.

\FloatBarrier
\section{Additional CLIP Results}
\label{app:clip_results}

\subsection{Training Settings}
\label{app:clip_settings_tasks}

Main ViT-B/32 models use batch size 128, learning rate $5\times10^{-6}$,
weight decay 0.1, 500 warmup steps, and three epochs. Dual-text uses two
image--text losses; trimodal adds RGB--IR alignment; grayscale and RGB-duplicate
controls preserve the trimodal objective and optimizer. Full-scale trimodal
results use seeds 42, 3407, and 2026. The matched ViT-B/32 and ViT-L/14
capacity comparison uses batch size 48 and the same three seeds. Retrieval is
evaluated bidirectionally for IR--text, RGB--text, and RGB--IR pairs.

The supplementary CLIP analysis answers four questions before presenting the
complete retrieval matrices: whether the full-scale result is repeatable across
seeds, how backbone capacity changes the result at fixed exposure, whether the
learned correspondence transfers to synchronized sensor-captured pairs, and
whether the initialization remains useful under task-specific adaptation. The
two wide retrieval tables are deferred to the end of this section so that the
experimental interpretation is established before the dense directional
breakdowns are introduced.

\begin{table}[H]
\centering
\caption{\textbf{Full-scale trimodal robustness across pretraining seeds.}}
\label{tab:app_seed_robustness}
\setlength{\tabcolsep}{3.5pt}
\footnotesize
\begin{tabular*}{\columnwidth}{@{\extracolsep{\fill}}lccc@{}}
\toprule
\textbf{Seed} & \textbf{Text Mean} & \textbf{Pair Mean} & \textbf{Overall Mean} \\
\midrule
42   & 40.00 & 95.31 & 67.66 \\
3407 & 40.31 & 95.36 & 67.84 \\
2026 & 40.39 & 95.38 & 67.88 \\
\midrule
\textbf{Mean $\pm$ SD} & $\mathbf{40.23\pm0.20}$ & $\mathbf{95.35\pm0.04}$ & $\mathbf{67.79\pm0.12}$ \\
\bottomrule
\end{tabular*}
\end{table}

Table~\ref{tab:app_seed_robustness} tests whether full-scale trimodal training
depends on a particular initialization. Text Mean spans only 0.39 points across
the three runs, Pair Mean spans 0.07 points, and Overall Mean spans 0.22 points.
The especially small Pair Mean variation indicates that direct paired-view
alignment is stable, while the slightly larger language-retrieval variation
does not change the ordering or the main conclusion. The reported three-seed
average therefore represents repeatable full-scale behavior rather than a
single favorable run.

\FloatBarrier
\begin{table}[H]
\centering
\caption{\textbf{Matched 50K backbone-capacity study.} Mean $\pm$ sample SD over three contrastive-training seeds. Directional R@1 cells list both retrieval directions in the order shown.}
\label{tab:app_capacity}
\fontsize{6.7}{8.1}\selectfont
\setlength{\tabcolsep}{2.2pt}
\renewcommand{\arraystretch}{1.12}
\begin{tabular*}{\columnwidth}{@{\extracolsep{\fill}}lcccc@{}}
\toprule
\textbf{Backbone} &
\makecell{\textbf{Text R@1}\\IR$\to$T / T$\to$IR} &
\makecell{\textbf{Text}\\\textbf{Mean}} &
\makecell{\textbf{Pair R@1}\\RGB$\to$IR / IR$\to$RGB} &
\makecell{\textbf{Pair}\\\textbf{Mean}} \\
\midrule
ViT-B/32 &
\makecell{$10.47\pm0.30$ /\\$10.43\pm0.07$} &
$24.06\pm0.02$ &
\makecell{$76.18\pm0.31$ /\\$76.31\pm0.16$} &
$86.22\pm0.15$ \\
ViT-L/14 &
\makecell{$16.39\pm0.07$ /\\$15.97\pm0.10$} &
$\mathbf{32.84\pm0.07}$ &
\makecell{$81.74\pm0.14$ /\\$83.74\pm0.44$} &
$\mathbf{90.35\pm0.17}$ \\
\bottomrule
\end{tabular*}
\end{table}

Table~\ref{tab:app_capacity} isolates backbone capacity from data scale by
holding the 50K training subset, objective, batch size, and three-seed protocol
fixed. ViT-L/14 raises Text Mean from 24.06 to 32.84 and Pair Mean from 86.22 to
90.35, with gains in both retrieval directions rather than in only one query
side. The larger improvement in Text Mean shows that additional visual capacity
is particularly useful for associating infrared-style structure with language,
whereas paired-view correspondence is already strong at 50K. This result
complements the main data-scaling ablation: dataset scale and encoder capacity
provide distinct, compatible sources of improvement.

\setcounter{table}{13}
\begin{table*}[!b]
\centering
\caption{\textbf{Per-source Text Mean / Pair Mean on unique test representatives.} Counts after joint-group deduplication are shown in the header.}
\label{tab:app_per_source_retrieval}
\setlength{\tabcolsep}{7pt}
\small
\renewcommand{\arraystretch}{1.08}
\begin{tabular*}{0.92\textwidth}{@{\extracolsep{\fill}}lccccc@{}}
\toprule
\textbf{Setting} & \textbf{NWPU (536)} & \textbf{RS5M (7,660)} & \textbf{RSICD (707)} & \textbf{RSITMD (97)} & \textbf{SkyScript (553)} \\
\midrule
OpenAI CLIP 0K & 9.58 / 11.97 & 4.36 / 11.54 & 5.52 / 11.17 & 27.32 / 38.14 & 5.21 / 9.22 \\
RemoteCLIP 0K & 8.92 / 15.02 & 0.75 / 5.46 & 9.74 / 15.94 & 36.77 / 45.53 & 3.22 / 11.30 \\
GeoRSCLIP 0K & 18.44 / 22.42 & 7.23 / 14.55 & 9.59 / 23.31 & 41.92 / 56.01 & 7.26 / 15.55 \\
RGB-only 580K & 18.10 / 18.91 & 11.97 / 21.90 & 11.55 / 18.86 & 35.91 / 52.23 & 11.15 / 16.49 \\
RGB duplicate 580K & 18.28 / 19.68 & 11.98 / 22.03 & 11.65 / 19.52 & 35.57 / 52.23 & 11.21 / 16.49 \\
Grayscale 580K & 22.67 / 30.53 & 16.02 / 31.92 & 13.81 / 35.17 & 41.58 / 67.18 & 13.38 / 23.93 \\
Dual-text 580K & \textbf{49.07} / 63.03 & \textbf{44.08} / 71.92 & 29.44 / 61.79 & 63.40 / 88.14 & \textbf{32.70} / 46.62 \\
Translated trimodal 50K & 33.55 / 86.63 & 25.71 / 88.52 & 18.58 / 86.82 & 52.06 / 94.85 & 18.78 / 79.11 \\
Translated trimodal 100K & 38.28 / 90.14 & 29.94 / 91.51 & 20.70 / 88.94 & 55.50 / 96.39 & 22.21 / 84.12 \\
Translated trimodal 300K & 43.84 / 93.22 & 38.32 / 94.83 & 26.26 / 92.31 & 59.62 / 97.42 & 27.34 / 90.27 \\
Translated trimodal 580K & 48.60 / \textbf{94.40} & 43.20 / \textbf{96.36} & \textbf{29.77} / \textbf{93.92} & \textbf{63.57} / \textbf{98.45} & 32.13 / \textbf{92.34} \\
\bottomrule
\end{tabular*}
\end{table*}
\setcounter{table}{10}

\begin{table}[H]
\centering
\caption{\textbf{Zero-shot exact-pair retrieval on synchronized Caltech Aerial RGB--thermal data.} Directional cells list R@1/R@5/R@10; Mean Recall averages all six values.}
\label{tab:app_caltech_pair}
\fontsize{6.6}{7.9}\selectfont
\setlength{\tabcolsep}{2.2pt}
\renewcommand{\arraystretch}{1.12}
\begin{tabular}{@{}p{0.31\columnwidth}C{0.23\columnwidth}C{0.23\columnwidth}c@{}}
\toprule
\textbf{Initialization / training} &
\makecell{\textbf{RGB$\to$T}\\\textbf{R@1/5/10}} &
\makecell{\textbf{T$\to$RGB}\\\textbf{R@1/5/10}} &
\makecell{\textbf{Mean}\\\textbf{Recall}} \\
\midrule
OpenAI CLIP~\cite{radford2021learning} & 7.89/29.39/39.43 & 7.89/21.15/31.54 & 22.88 \\
RemoteCLIP~\cite{liu2024remoteclip} & 7.17/18.28/25.09 & 5.38/13.26/22.58 & 15.29 \\
GeoRSCLIP~\cite{zhang2024rs5m} & 13.26/34.41/45.52 & 7.89/20.07/27.60 & 24.79 \\
\midrule
Translated view, dual-text loss & 23.66/57.35/72.40 & 15.41/39.78/51.97 & 43.43 \\
Grayscale view, trimodal loss & 16.85/44.44/55.56 & 15.77/34.41/45.16 & 35.36 \\
RGB duplicate, trimodal loss & 15.41/40.50/53.76 & 11.83/31.54/45.52 & 33.09 \\
Translated view, trimodal loss & \textbf{40.86/77.42/89.25} & \textbf{39.07/73.48/88.17} & \textbf{68.04} \\
\bottomrule
\end{tabular}
\end{table}

Table~\ref{tab:app_caltech_pair} evaluates zero-shot correspondence on
synchronized, sensor-captured RGB--thermal pairs, without using Caltech labels
or pair supervision during FusionRS training. The translated trimodal model
improves every reported recall depth in both directions and reaches 68.04 Mean
Recall, compared with 43.43 for dual-text training and 24.79 for the strongest
off-the-shelf reference. The reciprocal RGB$\to$thermal and
thermal$\to$RGB gains show that the result is not driven by a single retrieval
direction. Together with the controlled grayscale and RGB-duplicate rows, the
comparison attributes the transfer advantage to translated-view supervision
and explicit cross-view alignment.

\begin{table}[H]
\centering
\caption{\textbf{Additional adapted real-sensor tasks.} Mean $\pm$ sample SD over downstream seeds 42, 3407, and 2026. Caltech 10\% uses multilabel-stratified real training labels.}
\label{tab:app_transfer_boundary}
\setlength{\tabcolsep}{3.5pt}
\small
\resizebox{\columnwidth}{!}{
\begin{tabular}{lccc}
\toprule
\textbf{Initialization / training} & \textbf{SIRST-V2 IoU} & \textbf{IRSTD-1K IoU} & \textbf{Caltech 10\% mAP} \\
\midrule
OpenAI CLIP~\cite{radford2021learning} & $65.28\pm0.90$ & $55.11\pm1.16$ & $90.02\pm3.14$ \\
FusionRS, trimodal 580K & $65.13\pm1.38$ & $54.59\pm2.20$ & $89.71\pm3.19$ \\
\bottomrule
\end{tabular}
}
\end{table}

Table~\ref{tab:app_transfer_boundary} extends the transfer study from exact-pair
retrieval to target-specific segmentation and limited-label classification.
Across SIRST-V2, IRSTD-1K, and Caltech 10\%, FusionRS initialization remains
competitive with the original OpenAI CLIP initialization under the same
downstream seeds and adaptation recipes. The close three-seed averages show
that the learned representation can be reused without destabilizing supervised
adaptation on authentic infrared data. This table therefore complements the
strong zero-shot correspondence result in Table~\ref{tab:app_caltech_pair} by
demonstrating compatibility with task-specific real-sensor learning.

\subsection{Complete Retrieval Breakdowns}
\label{app:complete_retrieval_breakdowns}

Table~\ref{tab:app_per_source_retrieval} tests whether the aggregate pattern is
shared across all five source datasets rather than being determined by RS5M.
At 580K, translated trimodal training achieves the highest Pair Mean for every
source, ranging from 92.34 on SkyScript to 98.45 on RSITMD. It also obtains the
highest Text Mean on RSICD and RSITMD and remains close to the dual-text
specialist on the other sources. The consistent paired-view result across
different scene distributions and caption styles shows that the learned
RGB--infrared-style correspondence generalizes beyond the dominant source,
while the text scores preserve expected source-specific linguistic variation.

Each source cell reports Text Mean first and Pair Mean second, so the two
objectives remain visible even for the smaller subsets. Both values increase
monotonically with trimodal scale for every source from 50K through 580K.
The trend is therefore not produced by averaging a gain on one collection with
a decline on another. The deduplicated counts in the header also expose the
different evaluation denominators, making the per-source evidence directly
traceable to the aggregate test set.

\newpage
\begin{table*}[t]
\centering
\caption{\textbf{Full controlled retrieval results.} Values are percentages on 9,553 unique joint-group test representatives. Text Mean averages the six source-caption recalls; Pair Mean averages the six RGB--IR-style recalls.}
\label{tab:app_full_retrieval}
\fontsize{7.2}{8.4}\selectfont
\setlength{\tabcolsep}{2.4pt}
\renewcommand{\arraystretch}{1.18}
\resizebox{\textwidth}{!}{
\begin{tabular}{ccccccccccccccc}
\toprule
\multirow{2}{*}{\textbf{Setting}} & \multicolumn{3}{c}{\textbf{IR$\to$Text}} & \multicolumn{3}{c}{\textbf{Text$\to$IR}} & \multirow{2}{*}{\textbf{Text Mean}} & \multicolumn{3}{c}{\textbf{RGB$\to$IR}} & \multicolumn{3}{c}{\textbf{IR$\to$RGB}} & \multirow{2}{*}{\textbf{Pair Mean}} \\
& R@1 & R@5 & R@10 & R@1 & R@5 & R@10 & & R@1 & R@5 & R@10 & R@1 & R@5 & R@10 & \\
\midrule
OpenAI CLIP 0K & 1.14 & 2.86 & 4.50 & 1.35 & 4.10 & 6.10 & 3.34 & 8.44 & 16.15 & 20.08 & 2.18 & 4.30 & 5.95 & 9.52 \\
RemoteCLIP 0K & 0.27 & 0.77 & 1.46 & 0.26 & 1.08 & 1.71 & 0.92 & 3.74 & 8.04 & 10.72 & 1.20 & 2.78 & 3.85 & 5.06 \\
GeoRSCLIP 0K & 1.26 & 4.03 & 6.23 & 3.33 & 8.51 & 12.01 & 5.89 & 12.28 & 22.98 & 28.82 & 2.14 & 4.90 & 6.63 & 12.96 \\
RGB-only 580K & 2.44 & 6.98 & 10.76 & 6.04 & 14.05 & 20.01 & 10.05 & 16.14 & 28.55 & 34.90 & 5.65 & 11.50 & 15.56 & 18.72 \\
RGB duplicate 580K & 2.52 & 6.99 & 10.52 & 6.05 & 14.17 & 19.91 & 10.03 & 16.19 & 28.78 & 35.21 & 5.49 & 11.70 & 15.74 & 18.85 \\
Grayscale 580K & 4.38 & 12.30 & 17.65 & 7.39 & 17.32 & 23.79 & 13.81 & 21.50 & 34.75 & 41.77 & 16.13 & 26.21 & 31.79 & 28.69 \\
Dual-text 580K & 20.98 & 45.19 & 56.64 & 20.83 & 44.29 & 56.45 & 40.73 & 56.90 & 74.19 & 80.30 & 50.66 & 66.76 & 72.98 & 66.97 \\
Translated trimodal 50K & 9.64 & 24.74 & 34.48 & 10.06 & 24.69 & 33.58 & 22.87 & 76.36 & 89.68 & 93.12 & 75.64 & 88.72 & 91.97 & 85.92 \\
Translated trimodal 100K & 12.43 & 29.78 & 40.45 & 12.08 & 29.27 & 39.23 & 27.21 & 80.95 & 92.50 & 95.35 & 81.05 & 92.23 & 95.00 & 89.51 \\
Translated trimodal 300K & 17.51 & 39.17 & 50.41 & 17.14 & 38.10 & 49.50 & 35.31 & 86.60 & 95.98 & 97.98 & 87.32 & 95.82 & 97.55 & 93.54 \\
Translated trimodal 580K & 20.29 & 44.45 & 56.46 & 19.83 & 43.40 & 55.61 & 40.00 & 89.59 & 97.55 & 98.82 & 89.97 & 97.29 & 98.63 & \textbf{95.31} \\
\bottomrule
\end{tabular}
}
\end{table*}

Table~\ref{tab:app_full_retrieval} expands the compact retrieval summary in the
main paper into all twelve directional recall values. Across translated
trimodal training scales, every aggregate metric improves monotonically:
Text Mean rises from 22.87 at 50K to 40.00 at 580K, while Pair Mean increases
from 85.92 to 95.31. At full scale, RGB$\to$IR and IR$\to$RGB R@1 reach 89.59
and 89.97, respectively, compared with 56.90 and 50.66 under dual-text
training. Meanwhile, the two text-retrieval directions remain closely matched,
showing balanced query and target behavior. The full matrix therefore clarifies
that direct trimodal supervision chiefly strengthens paired-view
correspondence while retaining strong language retrieval.

The row groups make the ablation logic explicit. The 0K models measure
off-the-shelf transfer, RGB-only and RGB-duplicate training control for data
exposure and the number of visual inputs, grayscale controls for a
single-channel appearance change, and dual-text removes the direct
image--image term. The nested 50K--580K trimodal rows then vary only the amount
of translated training data. Reading the table in this order separates gains
from added examples, altered appearance, language supervision, and explicit
cross-view alignment.

\setcounter{table}{14}
\FloatBarrier
\section{Formal IR-Only Caption and VQA Protocol}
\label{app:vlm_results}

The formal evaluation uses a source-balanced set of 500 held-out test images.
Reference annotators inspect the aligned RGB--infrared-style pair, while every
evaluated model receives only the infrared-style image. Each final question is
answerable from that image alone. The release records the AI-assisted reference
provenance.

\subsection{Annotation and Quality Gates}

Three independent GPT-5.6 annotators (Sol, Terra, and Luna) inspect the same
aligned RGB--infrared-style pair in parallel. Caption references cover scene
semantics and observable infrared-style cues. VQA items cover scene category,
object presence, intensity, contrast, and spatial structure. Canonical items
have clear references, concise answers, and direct infrared-style evidence.

\begin{table}[H]
\centering
\caption{\textbf{Construction of the 500-image generative benchmark.}}
\label{tab:app_vlm_gates}
\fontsize{7.0}{8.4}\selectfont
\setlength{\tabcolsep}{2.4pt}
\renewcommand{\arraystretch}{1.12}
\begin{tabular}{@{}p{0.20\columnwidth}p{0.34\columnwidth}p{0.39\columnwidth}@{}}
\toprule
\textbf{Component} & \textbf{Target construction} & \textbf{Required gate} \\
\midrule
Images & 500 held-out images; 100 per source & Group-isolated test selection \\
Caption & Two tasks per image; up to three independent references per task & Valid reference for every image--task pair \\
VQA & Four question types per image & Canonical answer grounded in the IR image \\
Annotation & Three parallel GPT-5.6 annotators inspect RGB+IR & AI-assisted disclosure and retained provenance \\
Evaluation input & IR-style image and task prompt & Identical input format for all models \\
Scoring & One shared manifest and scorer & Identical items and metrics for all models \\
\bottomrule
\end{tabular}
\end{table}

Table~\ref{tab:app_vlm_gates} summarizes the controls applied before scoring.
Group isolation separates the benchmark from training data, the IR-answerability
gate removes questions that require RGB evidence, and the shared manifest keeps
prompts, references, and denominators identical across evaluated models. The
100-image allocation per source prevents the largest corpus from determining
the generative benchmark, while two caption tasks and four VQA types provide
coverage of both scene semantics and infrared-style evidence. Retaining
AI-assisted annotation provenance makes every accepted item auditable without
changing the IR-only input available to evaluated models. These controls ensure
that model comparisons use the same questions and references rather than
method-specific subsets.

\subsection{Tasks and Metrics}

\begin{table}[H]
\centering
\caption{\textbf{Caption/VQA tasks and metrics.} Results are additionally reported by source and task type.}
\label{tab:app_vlm_tasks}
\fontsize{6.8}{8.2}\selectfont
\setlength{\tabcolsep}{2.0pt}
\renewcommand{\arraystretch}{1.12}
\begin{tabular}{@{}p{0.18\columnwidth}p{0.28\columnwidth}p{0.23\columnwidth}p{0.24\columnwidth}@{}}
\toprule
\textbf{Task} & \textbf{Required evidence} & \textbf{Output} & \textbf{Primary reporting} \\
\midrule
Semantic caption & Objects and scene visible in IR & Short free-form description & ROUGE-L and content-word lexical F1 \\
IR-cue caption & Intensity, contrast, texture, and structure visible in IR & IR-grounded description & ROUGE-L, lexical F1, and IR-cue coverage \\
Scene/object VQA & Scene category or visible object evidence & Canonical short answer & Exact match and token F1 by type and source \\
Intensity/spatial VQA & Relative intensity, contrast, count, or layout evidence & Canonical short answer & Exact match and token F1 by type and source \\
\bottomrule
\end{tabular}
\end{table}

Table~\ref{tab:app_vlm_tasks} separates two generation objectives from two VQA
families. ROUGE-L and lexical F1 measure caption agreement, IR-cue coverage
checks whether observable modality cues are expressed, and exact match with
token F1 accommodates both canonical and partially matching VQA responses.
Semantic captions test preservation of objects and scene identity, whereas
IR-cue captions test whether intensity, contrast, texture, and structural
evidence are verbalized under the appropriate instruction. The VQA split
similarly distinguishes category and object recognition from relative
intensity, counting, and layout reasoning. Reporting by source and question
type prevents an aggregate score from hiding which visual evidence a model
uses successfully.

\clearpage
Tables~\ref{tab:app_qualitative_examples_a}--\ref{tab:app_qualitative_examples_h}
present 40 group-isolated \texttt{rc4} test examples, with one sample from each
source in every part. Both caption fields are shortened only for layout; the
IR-aware text is condensed from accepted benchmark references describing
observable intensity, contrast, boundary, texture, or spatial structure in the
infrared-style view.

\newcommand{\FusionQualRow}[4]{%
#1 &
\includegraphics[width=\linewidth]{figures/appendix_qualitative_rc4/#2_rgb.jpg} &
\includegraphics[width=\linewidth]{figures/appendix_qualitative_rc4/#2_ir.jpg} &
\textbf{Orig.} #3 \newline
\textbf{IR-aware.} #4 \\}

\begin{table}[H]
\centering
\caption{\textbf{Qualitative RGB--infrared-style annotation examples, part 1.} One strict test sample is shown per source.}
\label{tab:app_qualitative_examples_a}
\fontsize{6.2}{7.4}\selectfont
\setlength{\tabcolsep}{1.5pt}
\renewcommand{\arraystretch}{1.08}
\begin{tabular}{@{}C{0.14\columnwidth}C{0.18\columnwidth}C{0.18\columnwidth}>{\raggedright\arraybackslash}m{0.43\columnwidth}@{}}
\toprule
\textbf{ID} & \textbf{RGB} & \textbf{IR-style} & \textbf{Original / IR-aware Caption} \\
\midrule
\FusionQualRow{NWPU baseball}{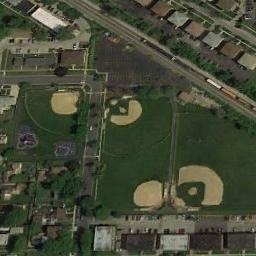}{Four baseball diamonds lie beside a parking lot and residential area.}{Repeated diamond-shaped outlines lie within broad fields, with bright linear roads along the margins.}
\midrule
\FusionQualRow{RS5M farmland}{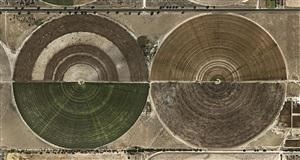}{An aerial view of several circular farm fields.}{Two large neighboring circular regions dominate the frame, each containing concentric ring patterns and darker borders.}
\midrule
\FusionQualRow{RSICD airport}{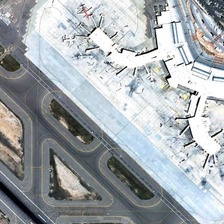}{Airport buildings and aircraft stand beside runways and open terrain.}{Bright aircraft-shaped outlines and angular pavement edges contrast with dark curving taxiways.}
\midrule
\FusionQualRow{RSITMD airport}{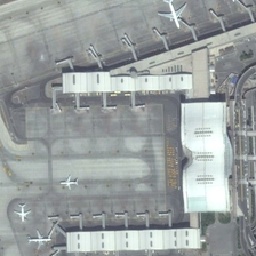}{Four planes are scattered near an airport terminal.}{Two long bright terminal-like bars border a dark apron containing repeated airplane-shaped outlines.}
\midrule
\FusionQualRow{SkyScript parking}{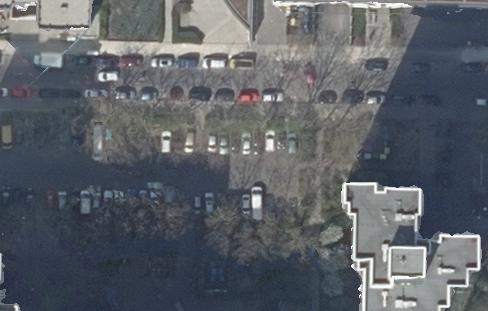}{A multi-storey building with a parking facility.}{Numerous small bright rectangles form rows across dark pavement, with angular building regions at the edges.}
\bottomrule
\end{tabular}
\end{table}

\begin{table}[H]
\centering
\caption{\textbf{Qualitative RGB--infrared-style annotation examples, part 2.} One strict test sample is shown per source.}
\label{tab:app_qualitative_examples_b}
\fontsize{6.2}{7.4}\selectfont
\setlength{\tabcolsep}{1.5pt}
\renewcommand{\arraystretch}{1.08}
\begin{tabular}{@{}C{0.14\columnwidth}C{0.18\columnwidth}C{0.18\columnwidth}>{\raggedright\arraybackslash}m{0.43\columnwidth}@{}}
\toprule
\textbf{ID} & \textbf{RGB} & \textbf{IR-style} & \textbf{Original / IR-aware Caption} \\
\midrule
\FusionQualRow{NWPU bridge}{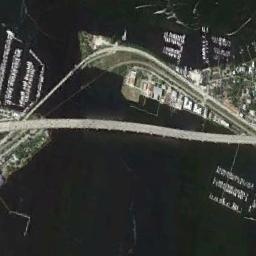}{Two bridges cross a river between harbor areas.}{A bright curving roadway and horizontal bridge sharply outline dark water, with compact bright structures along the shore.}
\midrule
\FusionQualRow{RS5M river}{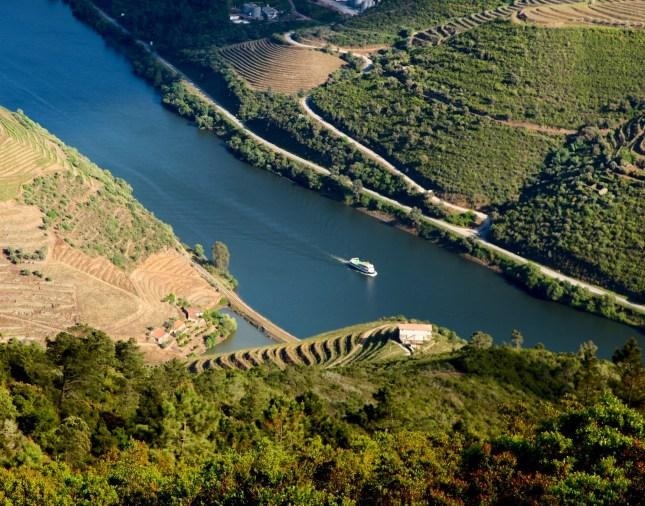}{A river flows through a valley and vineyards.}{A broad dark water channel runs diagonally between brighter textured slopes, with a small bright elongated object on the water.}
\midrule
\FusionQualRow{RSICD center}{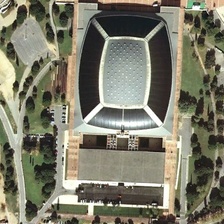}{Many trees surround a nearly rectangular central building.}{A bright rectangular complex surrounds a darker rounded enclosure, with mottled vegetation around its perimeter.}
\midrule
\FusionQualRow{RSITMD baseball}{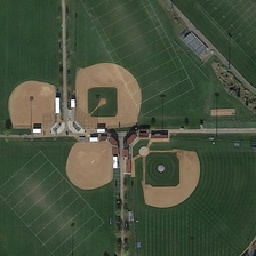}{Four baseball fields are near a football field.}{Four rounded field patterns radiate from a bright central intersection, with pale infields contrasting against dark outfields.}
\midrule
\FusionQualRow{SkyScript trailer park}{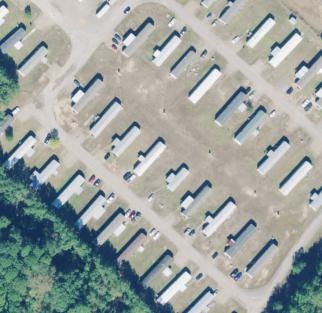}{An aerial image of a trailer park.}{Repeated bright elongated rectangles form diagonal rows separated by dark curving lanes.}
\bottomrule
\end{tabular}
\end{table}

\begin{table}[H]
\centering
\caption{\textbf{Qualitative RGB--infrared-style annotation examples, part 3.} One strict test sample is shown per source.}
\label{tab:app_qualitative_examples_c}
\fontsize{6.2}{7.4}\selectfont
\setlength{\tabcolsep}{1.5pt}
\renewcommand{\arraystretch}{1.08}
\begin{tabular}{@{}C{0.14\columnwidth}C{0.18\columnwidth}C{0.18\columnwidth}>{\raggedright\arraybackslash}m{0.43\columnwidth}@{}}
\toprule
\textbf{ID} & \textbf{RGB} & \textbf{IR-style} & \textbf{Original / IR-aware Caption} \\
\midrule
\FusionQualRow{NWPU residential}{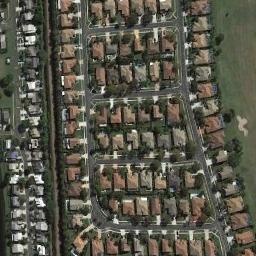}{A dense residential area contains neatly arranged houses and trees beside a lawn.}{Repeated bright-edged dark rectangles form rows along curved and straight linear corridors.}
\midrule
\FusionQualRow{RS5M city}{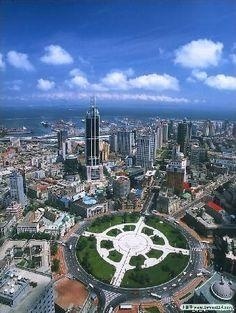}{An aerial view of a city with many buildings and a circular park.}{Bright building forms, roads, and a circular foreground structure contrast with the dark horizon.}
\midrule
\FusionQualRow{RSICD port}{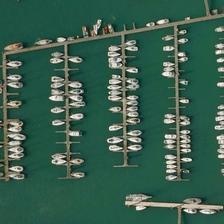}{Five rows of ships are ordered in a port.}{Bright boat-shaped marks cluster along thin parallel pier lines over a dark uniform background.}
\midrule
\FusionQualRow{RSITMD circular building}{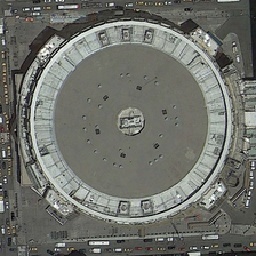}{Four roads surround the circular building.}{A circular ring with repeated dark radial segments encloses a smooth bright interior and a small central structure.}
\midrule
\FusionQualRow{SkyScript open pit}{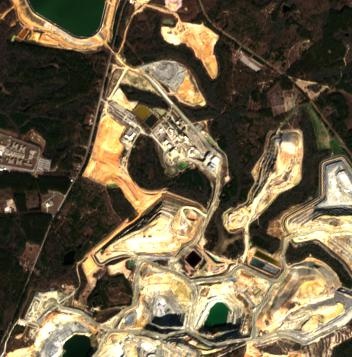}{A large open pit contains a body of water.}{Bright curving excavation edges enclose irregular dark basins, including a large dark water body at upper left.}
\bottomrule
\end{tabular}
\end{table}

\begin{table}[H]
\centering
\caption{\textbf{Qualitative RGB--infrared-style annotation examples, part 4.} One strict test sample is shown per source.}
\label{tab:app_qualitative_examples_d}
\fontsize{6.2}{7.4}\selectfont
\setlength{\tabcolsep}{1.5pt}
\renewcommand{\arraystretch}{1.08}
\begin{tabular}{@{}C{0.14\columnwidth}C{0.18\columnwidth}C{0.18\columnwidth}>{\raggedright\arraybackslash}m{0.43\columnwidth}@{}}
\toprule
\textbf{ID} & \textbf{RGB} & \textbf{IR-style} & \textbf{Original / IR-aware Caption} \\
\midrule
\FusionQualRow{NWPU harbor}{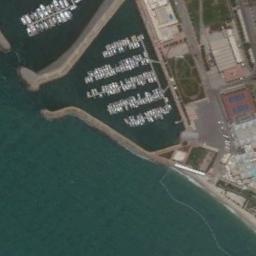}{A harbor with docked boats lies beside buildings, lawns, and courts.}{Dark water fills most of the frame, crossed by pale harbor edges and rows of bright vessel-shaped marks.}
\midrule
\FusionQualRow{RS5M power plant}{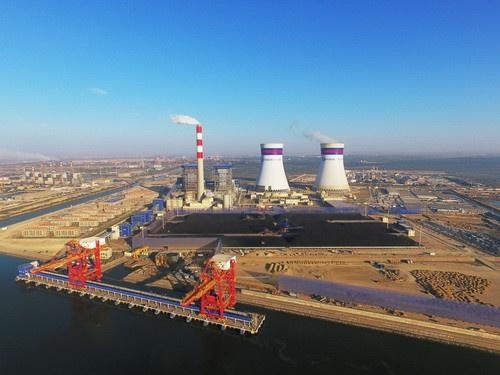}{An aerial view of a power plant with cooling towers.}{Two bright flared towers and a narrow vertical stack rise above angular industrial structures beside dark water.}
\midrule
\FusionQualRow{RSICD railway station}{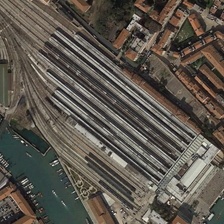}{The railway station is surrounded by residential areas.}{Numerous narrow alternating bright and dark lines run diagonally beside a broad dark region and bright-edged structures.}
\midrule
\FusionQualRow{RSITMD residential}{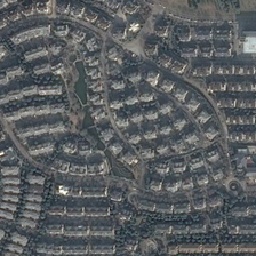}{A small wasteland area lies beside a residential district.}{Dense repeated bright-edged buildings line dark curving corridors across a highly textured scene.}
\midrule
\FusionQualRow{SkyScript farmland}{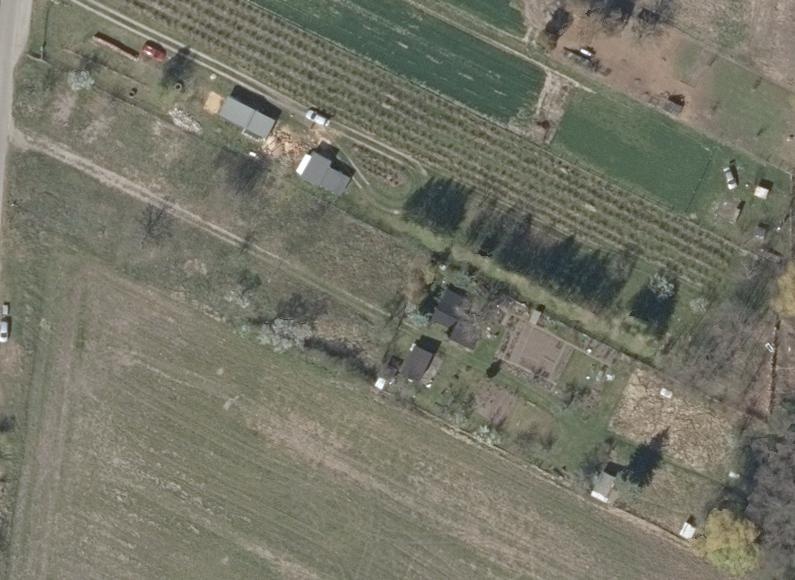}{A rural farm lies beside railway tracks.}{Repeated parallel bands cross the upper fields, with rectangular structures near the center and broad textured plots around them.}
\bottomrule
\end{tabular}
\end{table}

\begin{table}[H]
\centering
\caption{\textbf{Qualitative RGB--infrared-style annotation examples, part 5.} One strict test sample is shown per source.}
\label{tab:app_qualitative_examples_e}
\fontsize{6.2}{7.4}\selectfont
\setlength{\tabcolsep}{1.5pt}
\renewcommand{\arraystretch}{1.08}
\begin{tabular}{@{}C{0.14\columnwidth}C{0.18\columnwidth}C{0.18\columnwidth}>{\raggedright\arraybackslash}m{0.43\columnwidth}@{}}
\toprule
\textbf{ID} & \textbf{RGB} & \textbf{IR-style} & \textbf{Original / IR-aware Caption} \\
\midrule
\FusionQualRow{NWPU island}{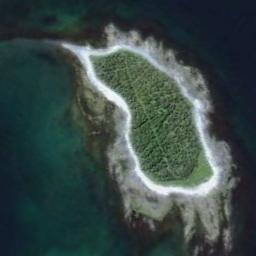}{An island contains dense vegetation and a reef.}{A dark textured oval interior is enclosed by a bright irregular rim against smoother surroundings.}
\midrule
\FusionQualRow{RS5M dam}{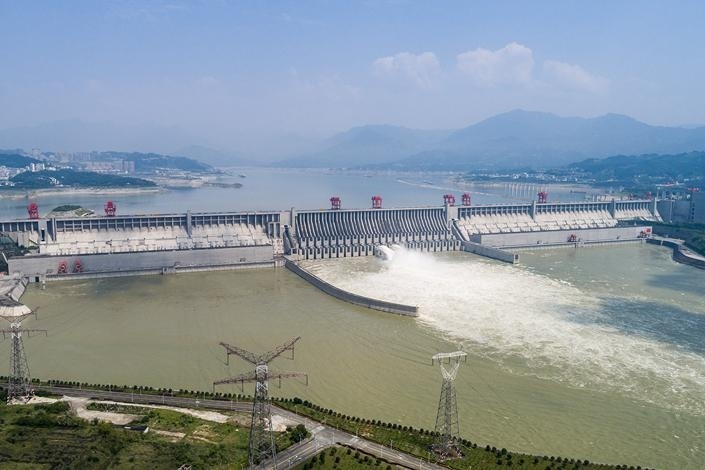}{An aerial view of a dam and surrounding water.}{A long bright horizontal barrier crosses a broad pale water surface, while dark mountains and sky form layered background bands.}
\midrule
\FusionQualRow{RSICD storage tanks}{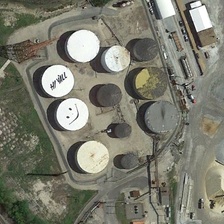}{Storage tanks are surrounded by meadows and bare land.}{Circular forms with varied grayscale interiors cluster beside a bright corridor and narrow rectangular structures.}
\midrule
\FusionQualRow{RSITMD factories}{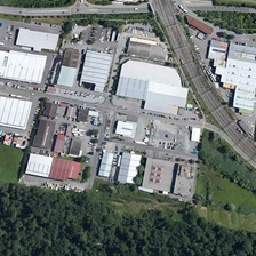}{Large factories stand on both sides of a road.}{Bright rectangular roofs and darker gaps form a dense built-up block beside a long linear corridor.}
\midrule
\FusionQualRow{SkyScript fields}{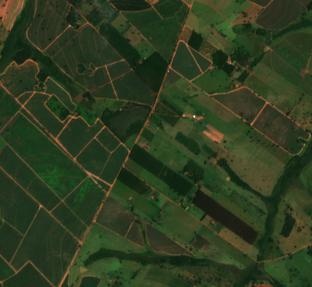}{A river runs through an agricultural field.}{Dark and light parcels form a mosaic separated by a dense network of thin bright boundaries.}
\bottomrule
\end{tabular}
\end{table}

\begin{table}[H]
\centering
\caption{\textbf{Qualitative RGB--infrared-style annotation examples, part 6.} One strict test sample is shown per source.}
\label{tab:app_qualitative_examples_f}
\fontsize{6.2}{7.4}\selectfont
\setlength{\tabcolsep}{1.5pt}
\renewcommand{\arraystretch}{1.08}
\begin{tabular}{@{}C{0.14\columnwidth}C{0.18\columnwidth}C{0.18\columnwidth}>{\raggedright\arraybackslash}m{0.43\columnwidth}@{}}
\toprule
\textbf{ID} & \textbf{RGB} & \textbf{IR-style} & \textbf{Original / IR-aware Caption} \\
\midrule
\FusionQualRow{NWPU overpass}{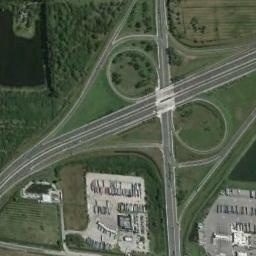}{A cloverleaf overpass crosses roads beside industrial areas.}{Several bright road corridors intersect diagonally, with two large loop-shaped ramps around a vertical route.}
\midrule
\FusionQualRow{RS5M intersection}{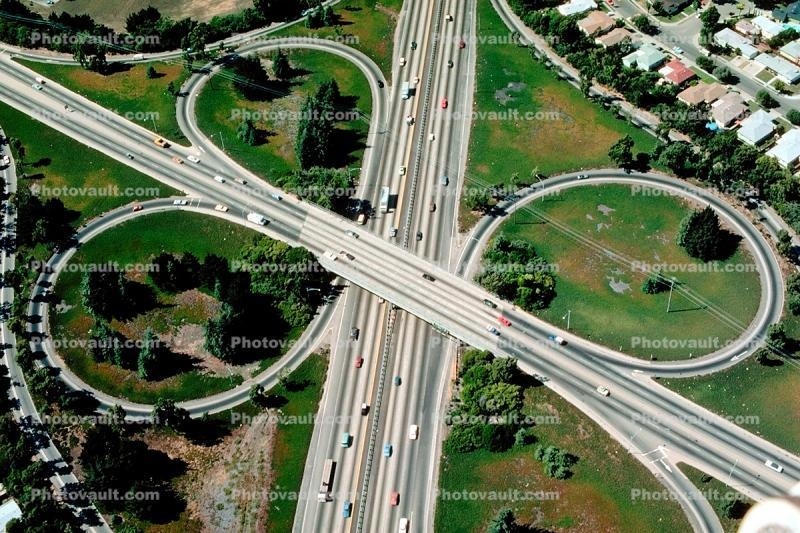}{An aerial view of a multilane highway intersection.}{Bright roads form a central cross with two curved loops, while small dark vehicle marks punctuate the lanes.}
\midrule
\FusionQualRow{RSICD viaduct}{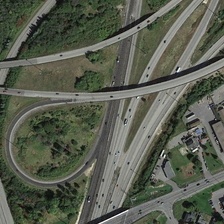}{A house with a gray roof stands beside an elevated roadway.}{Bright road surfaces cross at different angles, with a prominent curved loop and darker land between them.}
\midrule
\FusionQualRow{RSITMD lake}{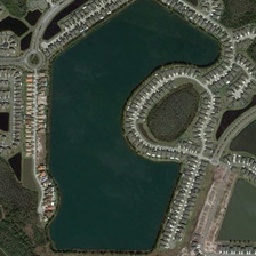}{Many buildings surround an irregular lake.}{A dark water body surrounds a lighter oval landform and bright curving routes with repeated shoreline structures.}
\midrule
\FusionQualRow{SkyScript forest}{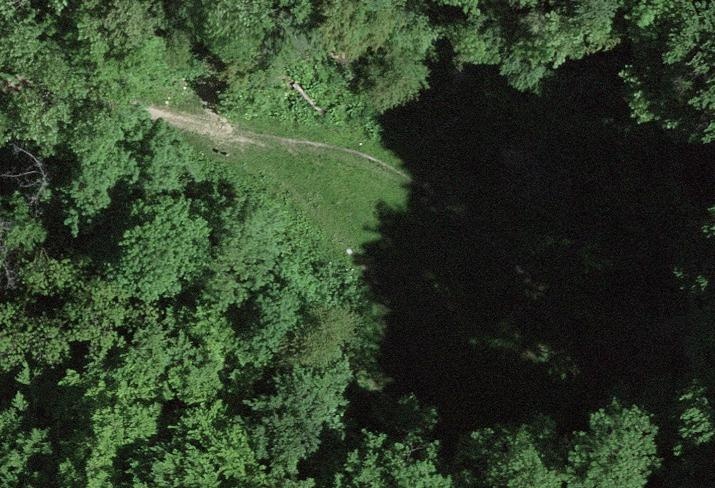}{A path runs through a forest.}{A narrow bright diagonal strip crosses textured foliage beside a broad dark open region.}
\bottomrule
\end{tabular}
\end{table}

\begin{table}[H]
\centering
\caption{\textbf{Qualitative RGB--infrared-style annotation examples, part 7.} One strict test sample is shown per source.}
\label{tab:app_qualitative_examples_g}
\fontsize{6.2}{7.4}\selectfont
\setlength{\tabcolsep}{1.5pt}
\renewcommand{\arraystretch}{1.08}
\begin{tabular}{@{}C{0.14\columnwidth}C{0.18\columnwidth}C{0.18\columnwidth}>{\raggedright\arraybackslash}m{0.43\columnwidth}@{}}
\toprule
\textbf{ID} & \textbf{RGB} & \textbf{IR-style} & \textbf{Original / IR-aware Caption} \\
\midrule
\FusionQualRow{NWPU roundabout}{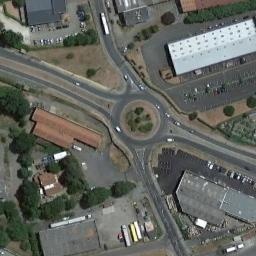}{A four-way roundabout is surrounded by buildings.}{A dark circular junction surrounds a lighter island, with radiating lanes and rectangular blocks at the edges.}
\midrule
\FusionQualRow{RS5M tennis}{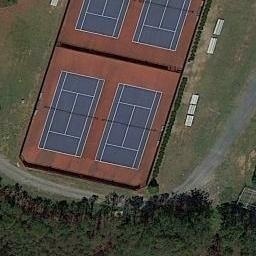}{The tennis courts are clearly visible in the aerial image.}{Four bright rectangular court interiors are separated by dark walkways and enclosed by a bright perimeter.}
\midrule
\FusionQualRow{RSICD river}{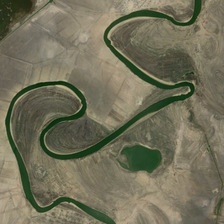}{A winding river crosses open terrain.}{A very dark sinuous channel loops across lighter terrain, with a separate dark oval feature nearby.}
\midrule
\FusionQualRow{RSITMD boats}{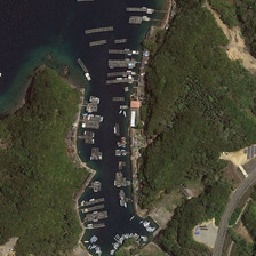}{Several boats lie within a narrow channel.}{A dark winding water channel is bordered by many bright elongated objects aligned along both banks.}
\midrule
\FusionQualRow{SkyScript tennis}{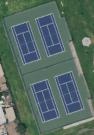}{An aerial view of tennis courts in a residential area.}{Four dark rectangles with repeated internal line patterns form a two-by-two arrangement within a bright area.}
\bottomrule
\end{tabular}
\end{table}

\begin{table}[H]
\centering
\caption{\textbf{Qualitative RGB--infrared-style annotation examples, part 8.} One strict test sample is shown per source.}
\label{tab:app_qualitative_examples_h}
\fontsize{6.2}{7.4}\selectfont
\setlength{\tabcolsep}{1.5pt}
\renewcommand{\arraystretch}{1.08}
\begin{tabular}{@{}C{0.14\columnwidth}C{0.18\columnwidth}C{0.18\columnwidth}>{\raggedright\arraybackslash}m{0.43\columnwidth}@{}}
\toprule
\textbf{ID} & \textbf{RGB} & \textbf{IR-style} & \textbf{Original / IR-aware Caption} \\
\midrule
\FusionQualRow{NWPU airplanes}{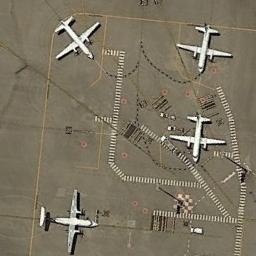}{Four planes are parked in different directions on an open apron.}{Four bright airplane outlines are distributed across a smooth gray apron, surrounded by repeated striped linear features.}
\midrule
\FusionQualRow{RS5M circular complex}{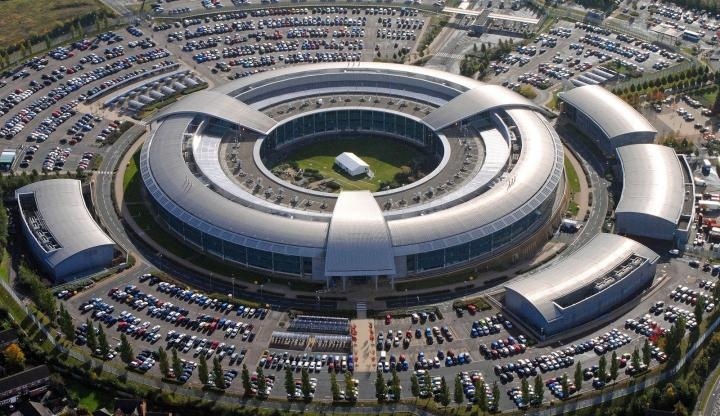}{A circular headquarters building is surrounded by parking lots.}{A bright ring-shaped building encloses a darker central courtyard and is surrounded by dense repeated vehicle-sized marks.}
\midrule
\FusionQualRow{RSICD residential}{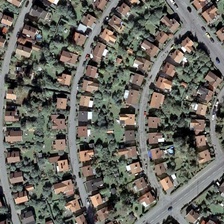}{Curved rows of houses enclose trees and lawns.}{Bright angular roofs form curved rows separated by smooth dark-gray roads and mottled vegetation.}
\midrule
\FusionQualRow{RSITMD stadium}{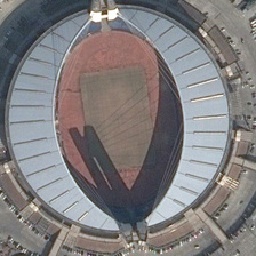}{Many cars and circular roads surround the circular stadium.}{A dark oval ring with repeated segments encloses a bright elongated central field and a darker diagonal shape.}
\midrule
\FusionQualRow{SkyScript ponds}{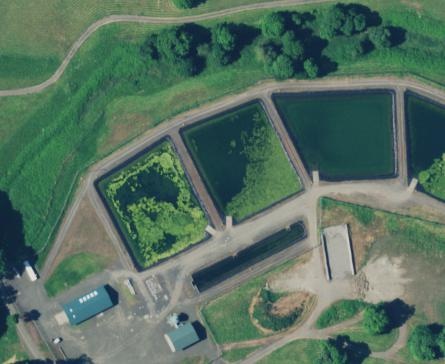}{A farm contains ponds and fenced areas.}{Dark polygonal basins are separated by bright straight embankments and bordered by mottled vegetation.}
\bottomrule
\end{tabular}
\end{table}

Together, these tables expose the annotation target beyond aggregate lexical
statistics. Scene identity remains recognizable in the source text, while the
IR-aware references consistently tie added language to observable grayscale
intensity, contrast, texture, boundaries, and spatial structure.

\end{document}